\documentclass{article}

\PassOptionsToPackage{round}{natbib}

\usepackage[preprint]{neurips_2020}

\usepackage[utf8]{inputenc} 
\usepackage[T1]{fontenc}    
\usepackage[colorlinks=true,linkcolor=red,citecolor=blue]{hyperref}       %
\usepackage{url}            
\usepackage{booktabs}       
\usepackage{amsfonts}       
\usepackage{nicefrac}       
\usepackage{microtype}      

\usepackage{algorithm}
\usepackage{algorithmic}

\usepackage{graphicx}
\usepackage{subfigure}
\usepackage{xcolor}

\usepackage{booktabs}

\usepackage{amsmath}
\usepackage{amsthm}
\usepackage{amssymb}
\usepackage{overpic}
\usepackage{enumitem}
\usepackage{multirow}
\usepackage{dsfont}
\usepackage{mathtools}
\usepackage{float}

\newcommand{\dataset}{\mathcal{D}}
\newcommand{\R}{\mathbb{R}}
\newcommand{\N}{\mathbb{N}}
\newcommand{\indi}{\mathds{1}}

\newcommand{\argmin}{\operatornamewithlimits{argmin}}
\newcommand{\Nrunf}{\text{\#\,Unf}}
\newcommand{\Maxviol}{\text{MVi}}
\newcommand{\sqcost}{\text{CoSq}}
\newcommand{\cost}{\text{Co}}
\newcommand{\Obj}{\text{Obj}}

\newcommand{\indf}{individually fair}
\newcommand{\indfn}{individual fairness}

\newcommand{\algorithmicreturn}{\textbf{return}}
\newcommand{\RETURN}{\STATE \algorithmicreturn}

\definecolor{mygreen}{rgb}{0,0.5,0}

\newtheorem{lemma}{Lemma}
\newtheorem{definition}{Definition}
\newtheorem{theorem}{Theorem}

\title{A Notion of Individual Fairness for Clustering}

\author{Matth{\"a}us Kleindessner \\
  University of Washington\\
  \texttt{mk1572@uw.edu} \\
   \And
   Pranjal Awasthi\\
   Rutgers University \& Google\\
   \texttt{pranjal.awasthi@rutgers.edu} \\
   \AND
  Jamie Morgenstern\\
   University of Washington \& Google \\
   \texttt{jamiemmt@cs.washington.edu}
}

\begin{document}

\maketitle

\begin{abstract}
  A common distinction in fair machine learning, in particular in fair
  classification, is between group fairness and individual fairness.  In the
  context of clustering, 
group fairness has been studied extensively 
in recent years; however, individual fairness for clustering has hardly been explored. 
In this paper, we propose~a~natural notion of individual fairness for clustering. 
Our notion asks that every data point, on average, is closer to the points in
its own cluster than to the points in any other cluster. 
We
study several questions related to our proposed notion of individual fairness.
On the negative side, we show that 
deciding whether a given data set allows for such an individually fair
clustering in general is NP-hard.  On the positive side, for the special case of
a 
data set lying on the real line, we propose an 
efficient dynamic programming approach to find an individually fair clustering.
For
general data sets, we investigate heuristics aimed at minimizing the number of
individual fairness violations and compare them
to standard clustering approaches on~real~data~sets.
\end{abstract}

\section{Introduction}\label{section_introduction}

Clustering is a classic unsupervised learning procedure 
and is
used in a wide range of fields to understand which data points are most
similar to each other, which regions in space a data set inhabits with
high density~\citep{ester1996density}, 
or 
to select representative
elements of a data set 
\citep{hastie2009}. The problem of clustering can be formulated 
in 
numerous 
ways, including objective-based formulations like
$k$-median \citep{awasthi2014center}, hierarchical
partitionings \citep{dasgupta2002performance}, and spectral clustering
\citep{Luxburg_tutorial}, which have also been considered subject to
additional constraints~\citep{wagstaff2001constrained}. A
recent surge in work has designed clustering algorithms to satisfy
various notions of proportional representation, including
proportionality for different demographics within
clusters~\citep{fair_clustering_Nips2017} or within the set of cluster
centers \citep{fair_k_center_2019}, or requiring a notion of
coherence on large subsets of a cluster~\citep{chen2019}.

All the latter  
proportionality constraints 
fall into the category of {\em group fairness}
constraints~\citep{friedler2016possibility}, which 
require a model to have similar statistical behavior for different
demographic groups. Such statistical guarantees necessarily give no
guarantee for any particular individual. For example, while profiles
of women might be equally represented in different clusters,  
such a clustering might not be a good clustering for any particular woman.
This weakness of proportionality constraints raises a natural
question: can one construct clusterings that provide fairness
guarantees for each individual,  
and what kind of fairness guarantees 
would an individual want to have after all?

We 
argue 
that if a clustering is used in a machine learning downstream task, then rather than  
caring about fairness of the clustering, one should care about fairness at the end of the pipeline and 
tune the clustering accordingly. This is analogous to using clustering as a preprocessing step for classification 
and caring about accuracy \citep{luxburg_ScienceOrArt}. However, if a clustering is used by a human decision maker, 
say for exploratory data analysis or resource allocation, an individual 
may strive for being well represented, which means  to be assigned to a cluster with similar data points. 
As a toy example, think of a company that clusters its customers and distributes semi-personalized coupons, where all customers 
in one cluster get the same coupons 
according to 
their 
(hypothesized)
preferences. 
A customer that ends up in a cluster with rather different other customers 
(and hence is not well represented by its cluster)
might get coupons that are less valuable to her than the 
coupons 
she would have got 
if she had been assigned to the 
cluster that is best representing her.

Motivated by such an example, 
our notion of individual fairness asks that each data point is assigned to the best representing cluster in the sense 
that the data
point, on average, is closer to the points~in its own
cluster than to the points in any other cluster.
 While our notion is related to a well-known concept of
clustering stability 
(cf. Section~\ref{sec_related_work_and_concepts}), 
many 
questions
are open. 
For instance, 
in contrast to the existing group fairness notions, 
an {\indf} clustering may not
exist (for a fixed number of clusters).
%
We make the following contributions towards
understanding 
{\indfn}~for~clustering:
\begin{itemize}[wide, labelwidth=!, labelindent=0pt]
\setlength{\itemsep}{-0.2pt}

\item We propose a natural notion of {\indfn} for clustering requiring 
that every 
data point, on average, is closer to the points in its own
cluster than to the points in any other cluster.
\item When the data lies on the real line, we show that an {\indf} clustering always
  exists, and we design an 
  efficient algorithm to find one.
  We 
  argue  why this 1-dim case 
  is interesting on its own.
\item We show that even for Euclidean data sets in $\R^2$,
  {\indf} clusterings might not exist 
  and 
  prove that the
  problem of deciding whether a given data set has an {\indf}
  $k$-clustering is NP-hard, even for $k=2$ 
  and when the underlying distance function is assumed to be a metric.
\item We perform experiments on real data sets and compare the performance of our
  polynomial time algorithm for the 1-dim case with $k$-means clustering. In the case of higher dimensions, we investigate several 
  standard clustering algorithms 
  with respect to our fairness notion.

\end{itemize}

\section{Fairness Notion}
Our notion of individual fairness 
applies to a data set~$\dataset$ together 
with a 
given  dissimilarity function~$d$ that measures how 
close
two data points are. 
We 
use the terms  dissimilarity and distance synonymously. 
We assume $d:\dataset\times \dataset\rightarrow\R_{\geq 0}$ to be 
symmetric with $d(x,x)=0$, but not necessarily to 
be a metric (i.e., to additionally satisfy the triangle~inequality and $d(x,y)=0 \Leftrightarrow x=y$).

Our fairness notion defines what it means that a data point is treated fair in a clustering of $\dataset$; namely: 
a data point is treated individually fair if the average distance to the points 
in its own cluster (the point itself excluded) is not greater than the average distance to the points in any other cluster. 
Then 
a clustering 
of $\mathcal{D}$ 
is said to be individually fair if it treats every data point 
of $\mathcal{D}$ 
individually fair.

\begin{figure}[t]
 \centering
 \includegraphics[scale=0.9]{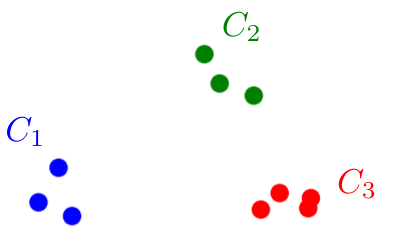}
 \hspace{3cm}
  \includegraphics[scale=0.9]{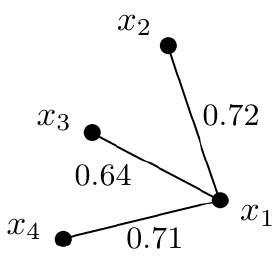}

 \caption{Two data sets in $\R^2$ with $d$ equaling the Euclidean metric. \textbf{Left:} An individually fair 3-clustering with clusters $C_1,C_2,C_3$.  
 \textbf{Right:} Four points for which there is no individually fair 2-clustering. For example, if 
 the two clusters were 
 $C_1=\{x_1,x_4\}$ and $C_2=\{x_2,x_3\}$, then $x_1$ would be treated unfair because of  
$d(x_1,x_4)=0.71>0.68=[d(x_1,x_2)+d(x_1,x_3)]/2$.
 }\label{figure_example}
\end{figure}

For the rest of the paper we assume $\dataset$ to be finite. Our definition of individual fairness for clustering can then be formally stated as follows 
(for $l\in\N$, we 
write 
$[l]=\{1,\ldots,l\}$):

\begin{definition}[Individually fair clustering]\label{def_indi_fairness}
 Let $\mathcal{C}=(C_1,\ldots,C_k)$ be a $k$-clustering of $\dataset$, that is 
 $\dataset=C_1\dot{\cup}\ldots\dot{\cup}C_k$ and 
 $C_i\neq \emptyset$ for $i\in[k]$. For $x\in \dataset$, we write $C(x)$ 
 for the cluster $C_i$ that $x$ belongs to. 
 We say that $x\in \dataset$ is 
treated individually fair
if either $C(x)=\{x\}$ or
 \begin{align}\label{def_individual_fairness_ineq}
\frac{1}{|C(x)|-1}
\sum_{y\in C(x)} 
d(x,y)\leq \frac{1}{|C_i|}\sum_{y\in C_i} d(x,y) 
 \end{align}
for all $i\in[k]$ with $C_i\neq C(x)$. The clustering $\mathcal{C}$ is 
individually fair
if every $x\in \dataset$ is treated individually fair.\footnote{For brevity, when it is clear from the context, instead of 
``individually fair'' we may only say ``fair''.}
\end{definition}

\begin{figure}[t]
\centering
\hspace{-6mm}\begin{overpic}[scale=0.24,trim=80 80 80 80,clip]{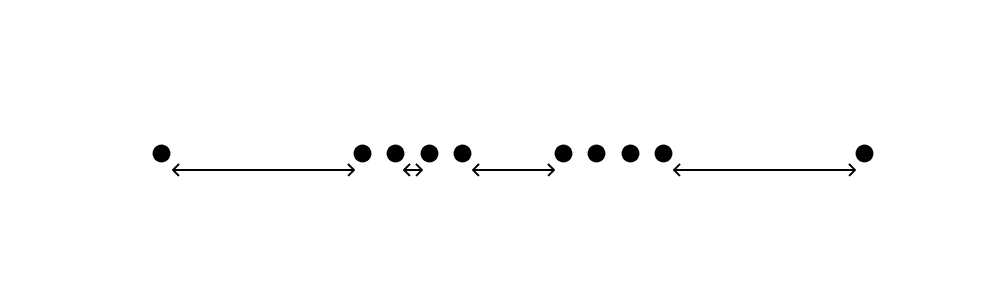}
 \put(21,1.2){$8$}
\put(51,1.2){$1$}
\put(38.3,0.6){\small $\frac{1}{3}$}
\put(81,1.2){$8$}
\put(108,12.5){\includegraphics[scale=0.25]{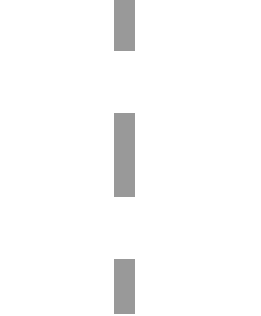}}
 \end{overpic}
 \hspace{2mm}
 \includegraphics[scale=0.24,trim=80 80 100 80,clip]{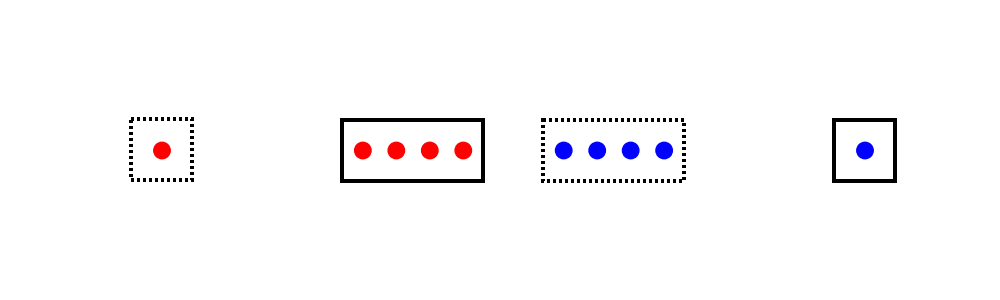}
 
 \caption{An example of a data set 
 on the real line with more than one individually fair clustering. \textbf{Left:} The data set and the distances between the points. 
 \textbf{Right:} The same data set with two 
 fair 
 2-clusterings (one encoded by color: red vs blue / one encoded by frames: solid vs dotted boundary).
 }\label{figure_example_2}
\end{figure}

We 
discuss some important observations about individually fair clusterings as defined in Definition~\ref{def_individual_fairness_ineq}: 
if in a clustering all clusters are well-separated and sufficiently far apart, 
then this clustering is 
fair. 
An example of such a scenario is provided in the left part of Figure~\ref{figure_example}. 
Hence, at least for such  simple clustering problems
with an ``obvious'' solution, 
individual fairness does not conflict with the clustering 
goal of
partitioning 
the data set  
such that 
``data points in the same cluster are similar to each other, and data points in different clusters are dissimilar'' \citep[][p.~306]{celebi2016}.
However, there are 
also 
data sets for which no 
fair $k$-clustering exists 
(for a fixed~$k$ 
and a given distance function~$d$).\footnote{
Of course, the trivial $1$-clustering $\mathcal{C}=(\dataset)$ or  
the 
trivial $|\dataset|$-clustering that puts every 
data 
point in a singleton are 
fair, 
and for a trivial distance function~$d\equiv 0$, every clustering is fair.
}
This can even happen for Euclidean data sets and $k=2$, as the right part of Figure~\ref{figure_example} shows.
If a data set allows 
for an individually fair $k$-clustering, there might be more than one fair $k$-clustering. An example of this is
shown in 
Figure~\ref{figure_example_2}. This example 
also 
illustrates 
that 
individual fairness does not necessarily work towards the 
aforementioned 
clustering 
goal.  
Indeed, in Figure~\ref{figure_example_2} the 
two 
clusters of the  clustering encoded by the frames, which is 
fair,  are not even~contiguous.  

These observations raise 
a 
number 
of questions such as: when does 
a 
fair $k$-clustering exist? Can we efficiently decide whether a fair $k$-clustering exists? 
If a fair $k$-clustering exists, can we efficiently compute it? Can we minimize some (clustering) objective 
over the set of all 
fair clusterings? If no fair $k$-clustering exists, can we find a clustering that violates inequality~\eqref{def_individual_fairness_ineq} only for a few data points, or 
a clustering that potentially violates \eqref{def_individual_fairness_ineq} for every data point, but only to a minimal extent? How do standard clustering algorithms such as 
Lloyd's algorithm (aka $k$-means) 
or linkage clustering \citep[e.g.,][Section 22]{shalev2014understanding}
perform in terms of 
fairness? Are there simple modifications to these 
algorithms 
in order to 
improve their~fairness?
In this paper, we explore some of these questions as outlined 
in Section~\ref{section_introduction}. 
%

\subsection{Related Work and Concepts}\label{sec_related_work_and_concepts}

We provide a detailed overview in Appendix~\ref{appendix_related_work}. Here we only present a brief summary.

\textbf{Existing Notions of Individual Fairness}~~~
\citet{fta2012} were the first to provide a notion of individual fairness by asking that 
similar data points (as measured by a given task-specific metric) should be treated
similarly by a randomized classifier. Subsequently, individual fairness has been studied in multi-armed bandit problems  
\citep{joseph2016,joseph2018,gillen2018}. 
The recent work of \citet{kearns_average_individual_fairness} 
introduces the notion of average individual fairness.

\textbf{Fairness for Clustering}~~~~The most established notion of fairness for clustering has been proposed by 
\citet{fair_clustering_Nips2017}. 
It asks that
each cluster has proportional representation from different
demographic groups. 
Several follow-up works extend 
that work 
\citep{roesner2018,sohler_kmeans,ahmadian2019,anagnostopoulos2019,backurs2019,bera2019,bercea2019,huang2019,fair_SC_2019,davidson2020}.

Alternative fairness notions for clustering 
are
tied to centroid-based clustering such as $k$-means, $k$-center and $k$-median
\citep{fair_k_center_2019,chen2019,jung2020}.
The recent notion of  
\citet{jung2020}  is the only one that comes with a guarantee for every single data point. It asks that every data point is 
somewhat close to a center, where ``somewhat'' depends on how close the data point is to its $k$ nearest neighbors 
and the motivation
for this notion 
comes from facility location.

\textbf{Average Attraction Property and Game-theoretic Interpretation}~~~~Our notion of individual fairness is closely related to the 
 average attraction property studied by \citet{vempala2008}, and our notion also has a game-theoretic interpretation.

\section{NP-Hardness}\label{section_np_hardness}
In this section we present one of the main results of our paper, stating the NP-hardness of deciding whether an individually fair $k$-clustering exists. 
For such a result, 
it is 
crucial 
to 
specify
how an input instance is encoded: we assume that a data set~$\dataset$ together with a distance function~$d$  
is 
represented by 
the distance matrix~$(d(x,y))_{x,y\in\dataset}$. 
Under this assumption we can prove  
the following~theorem:

\begin{theorem}[NP-hardness of individually fair clustering]\label{theorem_hardness}
Deciding whether a data set~$\dataset$ together with a distance function~$d$ has an individually fair $k$-clustering 
(for a given parameter $k$) 
is NP-hard. This even holds if $k=2$ is fixed and 
$d$ is required to be a metric.   
\end{theorem}

The proof of Theorem~\ref{theorem_hardness} is provided in Appendix~\ref{proof_hardness}. 
It shows NP-hardness 
of the individually fair clustering decision problem
via a reduction from a variant of 3-SAT. In this variant, we can assume a 3-SAT instance  
to have the 
same number of clauses as number of variables and that each variable occurs in at most three clauses. Given such a formula 
$\Phi=C_1 \wedge C_2\wedge \ldots \wedge C_n$ over variables $x_1,\ldots,x_n$, we construct a metric space~$(\dataset,d)$ with  
$\dataset=\{True, False,\star,\infty,C_1,\ldots,C_n,x_1,\neg x_1,\ldots,x_n,\neg x_n\}$ such that $\Phi$ is satisfiable if and only if 
$\dataset$ 
has an individually fair 2-clustering. 
The difficult part is in defining 
an appropriate metric 
$d$ 
to accomplish this.

Unless $\text{P}=\text{NP}$, Theorem~\ref{theorem_hardness} implies that 
for general data sets, even when being guaranteed that a fair $k$-clustering exists, 
there cannot be any efficient algorithm for computing such a fair clustering. 
However, as with all NP-hard problems, there are two possible remedies: first, we can restrict our considerations to data sets with some special structure. This is what we do 
in 
Section~\ref{section_1dim}, where we show that for 1-dimensional Euclidean data sets fair clusterings always exist and can be computed in polynomial time. 
We consider it to be 
an interesting question for follow-up work whether one can identify other classes of data sets with such a property (cf. Section~\ref{section_discussion}). 
Second, we can look at approximate versions of individual fairness in which we allow inequality~\eqref{def_individual_fairness_ineq} 
to be violated for a certain number of 
points or where we relax inequality~\eqref{def_individual_fairness_ineq} by introducing a multiplicative factor~$\gamma>1$ on its 
right 
side. 
We start exploring this direction in our experiments in Section~\ref{section_experiments_general}.

\section{1-dimensional Euclidean Case}\label{section_1dim}

One way to 
cope with the NP-hardness of the individually fair clustering problem is to 
restrict our considerations to data sets with some special structure. 
As an important example, 
here
we study the 
special 
case
of 
$\dataset\subseteq \R$ and $d$ being 
the 
Euclidean metric. 
We first show that in this case, for any $1\leq k\leq |\dataset|$, 
a 
fair $k$-clustering always exists.
In fact, 
we show that there 
exists 
a 
fair $k$-clustering with contiguous clusters. By contiguous clusters we mean that 
if $\dataset=\{x_1,\ldots,x_n\}$ with $x_1\leq x_2\leq \ldots\leq x_n$, 
the clustering is 
of the form $\mathcal{C}=(\{x_1,\ldots,x_{i_1}\},\{x_{i_1+1},\ldots,x_{i_2}\},\ldots,\{x_{i_{k-1}+1},\ldots,x_{n}\})$ 
for some $1\leq i_1<i_2<\ldots<i_{k-1}<n$. 
It might be surprising at a first glance that there also exist 
fair 
clusterings 
of 1-dimensional data sets 
with non-contiguous clusters, 
and indeed this seems to happen 
rarely, but it can happen as the example provided in Figure~\ref{figure_example_2} shows. 
%
Subsequently, 
we provide an 
efficient dynamic programming (DP) approach 
that finds 
a 
fair $k$-clustering solving
\begin{align}\label{1dim-problem}
 \min_{\substack{\mathcal{C}=(C_1,\ldots,C_k):~\mathcal{C}~\text{is a fair clus-}\\ \text{tering of $\dataset$ with contiguous clusters}}} \|(|C_1|-t_1,\ldots,|C_k|-t_k)\|_p,
\end{align}
where $t_1,\ldots,t_k\in[n]$ with $\sum_{i=1}^k t_i=n$ are 
given target cluster sizes, $p\in\R_{\geq 1}\cup\{\infty\}$ and $\|\cdot\|_p$ denotes the $p$-norm.

We believe that the results of this section are interesting on its own.
As 
an 
example consider the scenario that a teacher wants to give grades 
 based on the number of points that a student obtained 
 by setting some threshold values (e.g., a student gets a B 
if her number of points is in between 75 and~90). 
This can be interpreted as a 1-dim clustering problem, where clusters have to be contiguous and 
 individual fairness seems to be a highly desirable goal. Furthermore, 
 some teachers aim for a certain grade distribution (aka grading on a curve), in which case the problem 
 can 
 be phrased in the form of \eqref{1dim-problem}.
Clearly, one 
 can 
 think of similar examples in the context of credit scores or recidivism~risk~scores.

Let us now present our technical results 
(proofs  in Appendix~\ref{proof_existence}). 
A key observation is that a clustering with contiguous clusters is fair if and only if the boundary points of the clusters 
are treated fair:

\begin{lemma}[Fair boundary points imply fair clustering]\label{lemma_boundary_points}
 Let $\mathcal{C}=(C_1,\ldots,C_k)$ be a $k$-clustering of $\dataset=\{x_1,\ldots,x_n\}$, where $x_1\leq x_2\leq\ldots\leq x_n$, with contiguous clusters 
 $C_1=\{x_1,\ldots,x_{i_1}\},C_2=\{x_{i_1+1},\ldots,x_{i_2}\},\ldots,C_k=\{x_{i_{k-1}+1},\ldots,x_n\}$, for some 
 $1\leq i_1<\ldots<i_{k-1}<n$. Then 
 $\mathcal{C}$ is 
 individually 
 fair if and only if 
all points $x_{i_l}$ and $x_{i_l+1}$, $l\in[k-1]$,
 are treated 
 fair. 
 Furthermore, 
$x_{i_l}$ ($x_{i_l+1}$, resp.) is treated fair if and only if its average distance to the points in $C_l\setminus\{x_{i_l}\}$ ($C_{l+1}\setminus\{x_{i_l+1}\}$, resp.) 
is not greater than the average distance to the points in $C_{l+1}$ ($C_{l}$, resp.). 
\end{lemma}

The next theorem states that an individually fair $k$-clustering with contiguous clusters always exists.

\begin{theorem}[Existence of individually fair $k$-clustering]\label{lemma_existence_1D}
 Let  $\dataset\subseteq \R$ and $d$ be the Euclidean metric. 
 For any 
$k\in\{1,\ldots,|\dataset|\}$,  
 there exists an individually fair $k$-clustering of $\dataset$ with contiguous clusters.
\end{theorem}

The proof of Theorem~\ref{lemma_existence_1D} is constructive and provides an algorithm to compute a fair $k$-center 
clustering with contiguous clusters. This algorithm works by maintaining $k-1$ 
boundary indices, 
corresponding to a clustering
with contiguous clusters, 
and repeatedly increasing these 
indices until a fair 
clustering is found. We prove that at the 
latest when no 
index 
can be increased anymore, a fair clustering must have been found. 
However, the running time of the algorithm  scales exponentially with $k$. 

To overcome 
this, 
in the following we propose an efficient DP approach to find 
a 
solution~to~\eqref{1dim-problem}. 
Let 
$\dataset=\{x_1,\ldots,x_n\}$ with $x_1\leq \ldots\leq x_n$. 
Our approach builds a 
table~$T\in(\N\cup\{\infty\})^{n\times n\times k}$
with  
\begin{align}\label{definition_table_T}
T(i,j,l)=\min_{(C_1,\ldots,C_l)\in\mathcal{H}_{i,j,l}} \|(|C_1|-t_1,\ldots,|C_l|-t_l)\|_p^p
\end{align}
for $i\in[n]$, $j\in[n]$, $l\in[k]$, where
\begin{align*}
&\mathcal{H}_{i,j,l}=\big\{\mathcal{C}=(C_1,\ldots,C_l):~\text{$\mathcal{C}$ is a fair $l$-clustering of $\{x_1,\ldots,x_i\}$ with $l$ non-empty}\\
&~~~~~~~~~~~~~~~~~~~~~\text{contiguous clusters such that the right-most cluster $C_l$ contains exactly $j$ points}\big\}
\end{align*}
and $T(i,j,l)=\infty$ if $\mathcal{H}_{i,j,l}=\emptyset$. Here, we consider the case $p\neq \infty$. The modifications of our approach 
to 
the case~$p=\infty$ are minimal and 
are 
described in 
Appendix~\ref{appendix_p_equals_infty}. 

The optimal value of \eqref{1dim-problem} is given 
by $\min_{j\in[n]} T(n,j,k)^{1/p}$. Below, we will describe how to use the table~$T$ to compute an individually fair $k$-clustering solving \eqref{1dim-problem}. 
First, we explain how to build $T$. 
We have, for $i,j\in[n]$, 
\begin{align}\label{table_T_initial}
\begin{split}
 &T(i,j,1)=\begin{cases}
 |i-t_1|^p,& j=i, \\
 \infty, &j\neq i\\           
           \end{cases},
\qquad
T(i,j,i)=\begin{cases}
 \sum_{s=1}^i|1-t_s|^p,& j=1, \\
 \infty, &j\neq 1\\           
           \end{cases},\\
 &T(i,j,l)=\infty,\quad j+l-1>i,
 \end{split}
\end{align}
and
 the recurrence relation, for $l>1$ and $j+l-1\leq i$,
\begin{align}\label{recurrence_relation}
\begin{split}
T(i,j,l)=|j-t_l|^p+\min\left\{T(i-j,s,l-1):  s\in[i-j-(l-2)],\frac{\sum_{f=1}^{s-1}|x_{i-j}-x_{i-j-f}|}{s-1}\leq\right.\\
 \left.\frac{\sum_{f=1}^{j}|x_{i-j}-x_{i-j+f}|}{j},\frac{\sum_{f=2}^{j}|x_{i-j+1}-x_{i-j+f}|}{j-1}\leq\frac{\sum_{f=0}^{s-1}|x_{i-j+1}-x_{i-j-f}|}{s}\right\},
\end{split}
\end{align}
where we use the convention that $\frac{0}{0}=0$ for the 
fractions on the left sides of the inequalities. 
We explain the recurrence relation~\eqref{recurrence_relation} and argue why it is correct in Appendix~\ref{appendix_p_equals_infty}.

It is not hard to see that using~\eqref{recurrence_relation}, we can build the table~$T$ in 
time $\mathcal{O}(n^3k)$. Once we have $T$, we 
can compute a solution $(C_1^*,\ldots,C_k^*)$ to \eqref{1dim-problem} by specifying 
$|C_1^*|,\ldots,|C_k^*|$ 
in 
time $\mathcal{O}(nk)$ as follows: let $v^*=\min_{j\in[n]} T(n,j,k)$. We set $|C_k^*|=j_0$ for an arbitrary 
$j_0$ with $v^*=T(n,j_0,k)$. For $l=k-1,\ldots,2$, we then set $|C_l^*|=h_0$ for an arbitrary $h_0$~with 
(i) $T(n-\sum_{r=l+1}^k |C_r^*|,h_0,l)+\sum_{r=l+1}^k ||C_r^*|-t_r|^p=v^*$,
 (ii) the average distance of $x_{n-\sum_{r=l+1}^k |C_r^*|}$ to the closest $h_0-1$ many  points on its left side is not greater than the average 
distance to the points in $C_{l+1}^*$, and 
(iii) the average distance of $x_{n-\sum_{r=l+1}^k |C_r^*|+1}$ to the other points in $C_{l+1}^*$ is 
 not greater than the average distance to the closest $h_0$ many points on its left side. 
Finally, it is $|C_1^*|=n-\sum_{r=2}^k |C_r^*|$. It follows from the definition 
of the table~$T$ in~\eqref{definition_table_T} and Lemma~\ref{lemma_boundary_points} that for $l=k-1,\ldots,2$ we can always find some $h_0$ satisfying (i)~to~(iii)
and that our approach yields an individually fair $k$-clustering $(C_1^*,\ldots,C_k^*)$ of $\dataset$. 

Hence we have shown the following theorem:

\begin{theorem}[Efficient 
DP 
approach solves \eqref{1dim-problem}]
By means of the dynamic programming approach \eqref{definition_table_T} to \eqref{recurrence_relation} we can compute 
an individually 
fair clustering 
solving 
\eqref{1dim-problem} in running time $\mathcal{O}(n^3k)$.
\end{theorem}

\section{Experiments}\label{section_experiments}
We first study the case of 
1-dim Euclidean data, 
where 
we can 
apply 
our 
DP 
approach 
of
Section~\ref{section_1dim}.  
We then deal with general data sets. 
In this case, 
individually 
fair clusterings in the strict sense of Definition~\ref{def_indi_fairness}, 
which are required to treat every data point fair, may not  
exist, and 
even 
if they 
do, 
there is no efficient way to compute them (cf. Section~\ref{section_np_hardness}).    
Hence, we have to settle for approximate versions of Definition~\ref{def_indi_fairness} and fall back on approximation algorithms 
or 
heuristics.
As a starting~point~for~a~study of ``approximate individual fairness'' and a thorough search for 
approximation 
algorithms~with~guarantees 
(cf. Section~\ref{section_discussion}),~we~investigate the extent to which standard clustering algorithms 
violate 
individual fairness
and 
consider 
a heuristic approach 
for finding approximately fair~clusterings. 
Our experiments are intended to  serve as a proof of concept.
They do not focus on the running times of the 
algorithms or their applicability to \emph{large} data sets. Hence,  
we 
only 
use rather small data sets of sizes 500~to~1885.

Let us define some quantities: 
we measure the extent to which a $k$-clustering~$\mathcal{C}=(C_1,\ldots,C_k)$ of a dataset $\dataset$ is \mbox{(un-)fair} by 
$\Nrunf$ (``number unfair'') and $\Maxviol$ (``maximum violation'') defined as
\begin{align}\label{exp_quantities_unfair}
\Nrunf=|\{x\in \dataset: \text{$x$ is not treated fair}\}|,~\quad~ \Maxviol=\max_{x\in\dataset} \max_{C_i\neq C(x)}\frac{\frac{1}{|C(x)|-1}
\sum_{y\in C(x)} d(x,y)}{\frac{1}{|C_i|}\sum_{y\in C_i} d(x,y)},
\end{align}
where 
we use the convention that $\frac{0}{0}=0$. 
The clustering~$\mathcal{C}$ is fair if and only if
$\Nrunf=0$ and 
$\Maxviol\leq 1$. 
Mainly 
if $\dataset\subseteq \R^m$ and 
$d$ is the Euclidean metric, 
we measure the quality of~$\mathcal{C}$ 
(with respect to 
the 
goal of putting similar data points into the same cluster) by 
the $k$-means cost, referred to as $\sqcost$ (``cost squared''). In general, we measure the quality of~$\mathcal{C}$ by $\cost$ (``cost''),
which is 
compatible with 
Definition~\ref{def_indi_fairness} in that it uses ordinary 
rather than squared distances as $\sqcost$.  
It is 
\begin{align}\label{exp_cost_sq}
\sqcost =\sum_{i=1}^k \frac{1}{2 |C_i|}\sum_{x,y\in C_i}d(x,y)^2,\qquad  
 \cost =\sum_{i=1}^k \frac{1}{2|C_i|}\sum_{x,y\in C_i}d(x,y).
\end{align}
The reason for 
using $\sqcost$ as a measure of quality is 
to provide a fair evaluation of $k$-means clustering.

%
We performed all experiments  
in Python \textbf{(code in the supplementary material)}. 
We used the 
standard clustering algorithms from Scikit-learn or SciPy with all parameters set to their default~values.

\begin{table}[t]
\caption{Experiment on German credit data set. Clustering 1000 people according to their credit amount. Target cluster sizes $t_i=\frac{1000}{k}$, $i\in [k]$. 
\textsc{$k$-me++}$=k$-means++. 
Best values in~bold.
}
\label{experiment_1Da}

\begin{center}
\begin{small}
\begin{sc}
\begin{tabular}{lccccccccccc}
\toprule
 & $\Nrunf$ & $\Maxviol$ & $\Obj$ & $\sqcost$ & $\cost$ && $\Nrunf$ & $\Maxviol$ & $\Obj$ & $\sqcost$ & $\cost$ \\
\midrule
&  \multicolumn{5}{c}{$k=5$} &&  \multicolumn{5}{c}{$k=50$}\\
\addlinespace[0.1cm]
Naive &  105 & 2.95 & \textbf{0} & 4.78 & 23.53 && 101 & 2.6 & \textbf{0} & 0.19 & 3.06\\ 
DP & \textbf{0} & \textbf{1.0} & 172 & 1.39 & 17.62 && \textbf{0} & \textbf{1.0} & 8 & 0.08 & 2.29\\ 
$k$-means~ & 1 & 1.0 & 170 & 1.39 & \textbf{17.61} && 18 & 1.26 & 10 & 0.1 & 2.54\\ 
$k$-me++~& 0.79 & 1.0 & 279 & \textbf{1.38} & 19.36 && 11.04 & 1.15 & 50 & \textbf{0.01} & \textbf{1.72}\\
\bottomrule
\end{tabular}
\end{sc}
\end{small}
\end{center}
\vskip -0.1in
\end{table}

\subsection{
1-dimensional 
Euclidean 
Data Sets
}\label{section_experiments_1D}

We used the German Credit data set 
\citep{UCI_all_four_data_sets_vers2}.  
It 
comprises 1000 records (corresponding to 
human beings) 
and for each record one binary label (good vs. bad credit risk) and 20 features. 

In our first experiment, we clustered the 1000 people according to 
their credit amount, which is one of the 20 features.   
A~histogram of the data  
can be seen in 
Figure~\ref{plot_histogram_1D_data} in Appendix~\ref{appendix_exp_1dim}. We 
were aiming 
for $k$-clusterings with clusters of equal size (i.e., 
target cluster sizes $t_i=\frac{1000}{k}$, $i\in [k]$) and compared our DP approach of 
Section~\ref{section_1dim} with $p=\infty$ to 
$k$-means clustering 
as well as 
a naive clustering that simply puts the 
$t_1$ smallest 
points in the first cluster, the next $t_2$ many points in the second cluster, and so on. 
We considered two initialization strategies for $k$-means:  we either used the medians of the clusters of the naive clustering for initialization 
(thus, hopefully, biasing $k$-means towards the target cluster sizes) or we ran $k$-means++ \citep{kmeans_plusplus}. 
For the latter we report average 
results obtained from running the experiment for 100 times. In addition to the four quantities $\Nrunf$, $\Maxviol$, $\sqcost$ and  $\cost$ defined in 
\eqref{exp_quantities_unfair} and \eqref{exp_cost_sq}, we report $\Obj$ 
(``objective''),  
which is the value of the objective function of 
\eqref{1dim-problem} 
for $p=\infty$. 
Note that $k$-means 
yields contiguous clusters and 
$\Obj$ is meaningful for all four clustering methods that we consider.

The results are provided in Table~\ref{experiment_1Da} ($k=5$ and $k=50$) and in Table~\ref{experiment_1Da_part2} ($k=10$ and $k=20$) in Appendix~\ref{appendix_exp_1dim}. 
As expected, for the naive clustering we always have $\Obj=0$, 
for our  
DP approach (\textsc{DP}) we have $\Nrunf=0$ and 
$\Maxviol\leq 1$, and $k$-means++ (\textsc{$k$-me++}) performs best in terms of $\sqcost$. 
Most interesting to see is that both versions of $k$-means yield almost perfectly fair clusterings when $k$ is small 
and moderately fair clusterings when $k=50$ (with $k$-means++ outperforming $k$-means). 

In our second experiment (presented in Appendix~\ref{appendix_exp_1dim}), we used the first 500 records 
to train a multi-layer perceptron (MLP)
for predicting the label (good vs. bad credit risk). 
We then 
applied 
the MLP to estimate the probabilities of having a good credit risk for the other 500 people.    
We used the same clustering methods as in the first experiment to cluster 
the 500 people according to their probability estimate. 
We believe that such a clustering problem may arise 
frequently in practice (e.g., when a bank determines its lending policy) and that individual fairness is highly desirable in 
this 
context.


\subsection{General Data Sets}\label{section_experiments_general}

We performed the same set of experiments on the first 1000 records of the 
Adult data set, the Drug Consumption data set (1885 records), and the Indian Liver Patient data set (579 records) \citep{UCI_all_four_data_sets_vers2}. 
%
As distance function~$d$ we  used the Euclidean, 
Manhattan or Chebyshev metric.  
Here we only present the results for the Adult data set and the Euclidean metric, the other results  are provided in Appendix~\ref{appendix_exp_general}.
Our observations are largely consistent between the different data sets~and~metrics. 

\vspace{1mm}
\textbf{First Experiment --- (Un-)Fairness of Standard Algorithms}~~~~Working with  the Adult data set, 
we only used its six numerical features
(e.g., age, hours worked per week), normalized to zero mean
and unit variance, for representing records. 
We applied several standard clustering algorithms as well as the group-fair $k$-center algorithm of 
\citet{fair_k_center_2019} (referred to as $k$-center~GF) 
to the data set ($k$-means++; $k$-medoids; spectral clustering (SC)) or its 
distance matrix
($k$-center using the greedy strategy of \citet{gonzalez1985}; $k$-center GF; single / average / complete linkage clustering). 
%
In order to study the extent to which these methods produce (un-)fair clusterings, for $k=2,5,10,20,30,\ldots,100$, 
we computed 
$\Nrunf$ and $\Maxviol$ as defined in~\eqref{exp_quantities_unfair} for the resulting $k$-clusterings. 
For measuring the quality of the clusterings we computed $\sqcost$ or $\cost$ as defined in~\eqref{exp_cost_sq}.

\newcommand{\scaleA}{0.275}

\begin{figure*}[t]
\centering
%
%
%
\includegraphics[width=\textwidth]{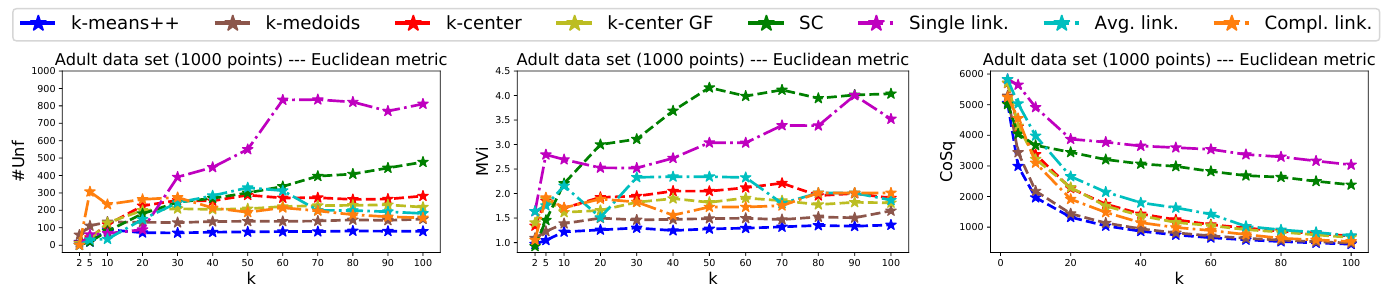}

\vspace{-2mm}
\caption{$\Nrunf$ 
\textbf{(left)}, $\Maxviol$ 
\textbf{(middle)} and $\sqcost$ \textbf{(right)}
for the clusterings produced by the 
various 
algorithms as a function of 
$k$.
}\label{exp_gen_standard_alg_Adult}
\end{figure*}

The results are provided in 
Figure~\ref{exp_gen_standard_alg_Adult}.  
For $k$-means++, $k$-medoids, $k$-center, $k$-center GF and SC we show average results obtained from running 
them 
for 
25 times since their outcomes depend on random initializations. 
We can see that, in particular for large values of~$k$, $k$-center, $k$-center GF, SC, and the linkage algorithms can be quite unfair 
with 
rather large values of $\Nrunf$ and $\Maxviol$. 
In contrast, $k$-means++ produces rather fair clusterings with $\Nrunf\leq 90$ and $\Maxviol\leq 1.4$ even when $k$ is large. 
For 
a 
baseline comparison, for a random clustering in which every data point was assigned to one of $k$ clusters uniformly at random we observed 
$\Nrunf= 990$ and $\Maxviol=4.0$ on average (when $k=100$). 
The beneficial behavior of $k$-means++ with respect to our notion of individual fairness 
raises the question whether one 
can 
prove guarantees on the extent 
to which clusterings produced by 
$k$-means++ are fair (cf. Section~\ref{section_discussion}). 
The $k$-medoids algorithm performs worse than $k$-means++, but better than the other algorithms. The clusterings produced by 
$k$-center~GF, which we ran with the constraint of choosing $\lfloor k/2\rfloor$ female and $\lceil k/2\rceil$ male centers,  
are slightly more fair than the ones produced by $k$-center.
However, note that it really depends on the data set whether~a group-fair 
clustering is individually fair or not (example 
provided 
in Appendix~\ref{example_group_fair_vs_indi_fair}). 
Unsurprisingly, $k$-means++ outperforms the other methods in terms of $\sqcost$ since it is designed with 
the~goal~of~minimizing~this~quantity.

\begin{figure*}[t]
\centering
%
\includegraphics[width=\textwidth]{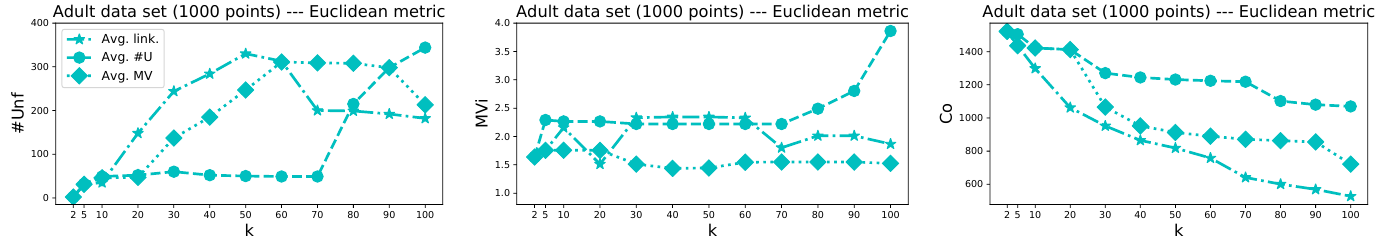}

\vspace{-2mm}
\caption{$\Nrunf$ 
\textbf{(left)}, $\Maxviol$ 
\textbf{(middle)} and $\cost$ \textbf{(right)}
for the clusterings produced by average linkage clustering 
and the two variants of our heuristic 
to improve it: the first ($\#$U in the legend) greedily chooses splits as to minimize $\Nrunf$, the second 
(MV) 
as to minimize $\Maxviol$.
}
\label{exp_gen_heuristics_Adult}
\end{figure*}

\vspace{1mm}
\textbf{Second Experiment --- Heuristics to Improve Linkage Clustering}~~~~One might wonder whether there are 
modifications to the standard clustering algorithms 
that 
make them more fair.  
A natural idea to make any clustering 
more fair is to make local changes to 
it 
and iteratively pick a data point that is not treated fair and assign 
it to the cluster that it is closest too. 
After picking and reassigning a data point, this 
point is treated fair. However, 
in 
experiments we observed that  usually we can only pick 
a very small number of data points whose reassignment does not cause other 
points that are initially treated fair to be treated unfair after the reassignment 
(example provided in Appendix~\ref{example_local_search}). 

Another idea that we want to present here is specifically tied to linkage clustering. 
As our experiments show this idea results in  linkage clustering producing clusterings that are significantly more fair than the ones produced by ordinary linkage clustering. 
Linkage clustering 
builds 
a binary tree that represents a hierarchical clustering with the root of the tree corresponding to the whole data set and 
every 
node 
corresponding to a subset such that a parent is the union of its two children.   
The leaves of the tree correspond to singletons comprising one data point 
\citep[e.g., ][Section 22.1]{shalev2014understanding}. If one wants to obtain a $k$-clustering of the data set, 
the output of a linkage clustering 
algorithm is a certain pruning of this tree.
When individual fairness is a goal, 
we propose to construct a $k$-clustering / a pruning of the tree as follows 
(pseudocode provided in Appendix~\ref{appendix_pruning_strategy}): starting with the two children of the root, 
we maintain a set of 
nodes 
that corresponds to a clustering 
and proceed in $k-2$ rounds. In round~$i$, we greedily 
split 
one of the $i+1$ many 
nodes 
that we currently have into its two children such that the resulting $(i+2)$-clustering minimizes, over the $i+1$ many possible splits, 
 $\Nrunf$ as defined in 
\eqref{exp_quantities_unfair}. Alternatively, we can split the node that 
gives rise to a minimum value of 
$\Maxviol$ (also defined in \eqref{exp_quantities_unfair}).

In Figure~\ref{exp_gen_heuristics_Adult}, we show $\Nrunf$, $\Maxviol$ and $\cost$ for ordinary average linkage clustering and a modified version 
using our heuristic approach in its both variants ($\#$U denotes the variant 
based on
$\Nrunf$ and MV the variant 
based on 
$\Maxviol$). 
Analogous experiments 
with single or complete instead of average linkage clustering are presented 
in Appendix~\ref{appendix_exp_general}. We can see that our 
approach leads to a 
significant 
improvement in $\Nrunf$ (for $k\leq 50$ this holds for both variants, 
but in particular for the variant 
aiming to minimize $\Nrunf$). The variant 
based on 
$\Maxviol$ leads to an improvement in $\Maxviol$. However, these improvements come at the price of an increase in 
$\cost$ as we can see from the right plot of Figure~\ref{exp_gen_heuristics_Adult}.

\section{Discussion}\label{section_discussion}
 In this work we 
contributed to 
the study of individual fairness in the context of clustering, which is only in its infancy. 
We proposed a notion of individual fairness that aims at data points being well represented by their clusters. 
Formally, it asks that every data
point, on average, is closer to the points in its own
cluster than to the points in any other cluster. 
This notion raises numerous questions, some of which we addressed: we showed that for general data sets, it is NP-hard to decide whether an individually fair $k$-clustering exists. 
In contrast, for one-dimensional Euclidean data sets we can compute a fair clustering by means of an efficient dynamic programming approach.
We examined standard clustering algorithms 
and saw that 
$k$-means++ 
often produces clusterings that are only slightly unfair. We also studied a simple heuristic to make linkage clustering more fair.

Still, many questions remain open, and we hope to inspire follow-up work to address some of these: using our measures $\Nrunf$ or $\Maxviol$ 
(cf. Section~\ref{section_experiments}), or some other measure, to define a notion of ``approximate individual fairness'', can we design 
algorithms with provable guarantees for finding such an approximately fair clustering? 
Can we do so for general data sets, or which assumptions about the data set do we need to make? 
Are there classes of data sets other than 1-dimensional Euclidean 
ones 
that 
allow for a (strictly) individually fair clustering? 
Finally, can we provide guarantees for Euclidean data sets and $k$-means++ clustering, which performed surprisingly well in our experiments? 

%
%


\bibliography{mybibfile_fairness}

\begin{thebibliography}{41}
\providecommand{\natexlab}[1]{#1}
\providecommand{\url}[1]{\texttt{#1}}
\expandafter\ifx\csname urlstyle\endcsname\relax
  \providecommand{\doi}[1]{doi: #1}\else
  \providecommand{\doi}{doi: \begingroup \urlstyle{rm}\Url}\fi

\bibitem[Ahmadian et~al.(2019)Ahmadian, Epasto, Kumar, and
  Mahdian]{ahmadian2019}
S.~Ahmadian, A.~Epasto, R.~Kumar, and M.~Mahdian.
\newblock Clustering without over-representation.
\newblock In \emph{ACM SIGKDD Conference on Knowledge Discovery and Data Mining
  (KDD)}, 2019.

\bibitem[Anagnostopoulos et~al.(2019)Anagnostopoulos, Becchetti, Böhm,
  Fazzone, Leonardi, Menghini, and Schwiegelshohn]{anagnostopoulos2019}
A.~Anagnostopoulos, L.~Becchetti, M.~Böhm, A.~Fazzone, S.~Leonardi,
  C.~Menghini, and C.~Schwiegelshohn.
\newblock Principal fairness: Removing bias via projections.
\newblock arXiv:1905.13651 [cs.DS], 2019.

\bibitem[Arthur and Vassilvitskii(2007)]{kmeans_plusplus}
D.~Arthur and S.~Vassilvitskii.
\newblock k-means++: The advantages of careful seeding.
\newblock In \emph{Symposium on Discrete Algorithms (SODA)}, 2007.

\bibitem[Awasthi and Balcan(2014)]{awasthi2014center}
P.~Awasthi and M.-F. Balcan.
\newblock Center based clustering: A foundational perspective.
\newblock In \emph{Handbook of Cluster Analysis}. CRC Press, 2014.

\bibitem[Backurs et~al.(2019)Backurs, Indyk, Onak, Schieber, Vakilian, and
  Wagner]{backurs2019}
A.~Backurs, P.~Indyk, K.~Onak, B.~Schieber, A.~Vakilian, and T.~Wagner.
\newblock Scalable fair clustering.
\newblock In \emph{International Conference on Machine Learning (ICML)}, 2019.

\bibitem[Balcan et~al.(2008)Balcan, Blum, and Vempala]{vempala2008}
M.-F. Balcan, A.~Blum, and S.~Vempala.
\newblock A discriminative framework for clustering via similarity functions.
\newblock In \emph{ACM Symposium on Theory of Computing (STOC)}, 2008.

\bibitem[Bera et~al.(2019)Bera, Chakrabarty, Flores, and Negahbani]{bera2019}
S.~Bera, D.~Chakrabarty, N.~Flores, and M.~Negahbani.
\newblock Fair algorithms for clustering.
\newblock In \emph{Neural Information Processing Systems (NeurIPS)}, 2019.

\bibitem[Bercea et~al.(2019)Bercea, Groß, Khuller, Kumar, Rösner, Schmidt,
  and Schmidt]{bercea2019}
I.~O. Bercea, M.~Groß, S.~Khuller, A.~Kumar, C.~Rösner, D.~R. Schmidt, and
  M.~Schmidt.
\newblock On the cost of essentially fair clusterings.
\newblock In \emph{Approximation, Randomization, and Combinatorial
  Optimization. Algorithms and Techniques (APPROX/RANDOM)}, 2019.

\bibitem[Celebi and Aydin(2016)]{celebi2016}
M.~E. Celebi and K.~Aydin.
\newblock \emph{Unsupervised Learning Algorithms}.
\newblock Springer, 2016.

\bibitem[Chen et~al.(2019)Chen, Fain, Lyu, and Munagala]{chen2019}
X.~Chen, B.~Fain, L.~Lyu, and K.~Munagala.
\newblock Proportionally fair clustering.
\newblock In \emph{International Conference on Machine Learning (ICML)}, 2019.

\bibitem[Chierichetti et~al.(2017)Chierichetti, Kumar, Lattanzi, and
  Vassilvitskii]{fair_clustering_Nips2017}
F.~Chierichetti, R.~Kumar, S.~Lattanzi, and S.~Vassilvitskii.
\newblock Fair clustering through fairlets.
\newblock In \emph{Neural Information Processing Systems (NIPS)}, 2017.

\bibitem[Dasgupta(2002)]{dasgupta2002performance}
S.~Dasgupta.
\newblock Performance guarantees for hierarchical clustering.
\newblock In \emph{International Conference on Computational Learning Theory
  (COLT)}, 2002.

\bibitem[Davidson and Ravi(2020)]{davidson2020}
I.~Davidson and S.~S. Ravi.
\newblock Making existing clusterings fairer: Algorithms, complexity results
  and insights.
\newblock In \emph{AAAI Conference on Artificial Intelligence}, 2020.

\bibitem[Dua and Graff(2019)]{UCI_all_four_data_sets_vers2}
D.~Dua and C.~Graff.
\newblock {UCI} machine learning repository, 2019.
\newblock German Credit data set available on
  \url{https://archive.ics.uci.edu/ml/datasets/Statlog+(German+Credit+Data)}.
  Adult data set available on
  \url{https://archive.ics.uci.edu/ml/datasets/adult}. Drug Consumption data
  set available on
  \url{https://archive.ics.uci.edu/ml/datasets/Drug+consumption+(quantified)}.
  Indian Liver Patient data set available on
  \url{https://archive.ics.uci.edu/ml/datasets/ILPD+(Indian+Liver+Patient+Dataset)}.

\bibitem[Dwork et~al.(2012)Dwork, Hardt, Pitassi, Reingold, and Zemel]{fta2012}
C.~Dwork, M.~Hardt, T.~Pitassi, O.~Reingold, and R.~Zemel.
\newblock Fairness through awareness.
\newblock In \emph{Innovations in Theoretical Computer Science Conference
  (ITCS)}, 2012.

\bibitem[Ester et~al.(1996)Ester, Kriegel, Sander, and Xu]{ester1996density}
M.~Ester, H.-P. Kriegel, J.~Sander, and X.~Xu.
\newblock A density-based algorithm for discovering clusters in large spatial
  databases with noise.
\newblock In \emph{International Conference on Knowledge Discovery and Data
  Mining (KDD)}, 1996.

\bibitem[Feldman et~al.(2015)Feldman, Friedler, Moeller, Scheidegger, and
  Venkatasubramanian]{feldman2015}
M.~Feldman, S.~A. Friedler, J.~Moeller, C.~Scheidegger, and
  S.~Venkatasubramanian.
\newblock Certifying and removing disparate impact.
\newblock In \emph{ACM International Conference on Knowledge Discovery and Data
  Mining (KDD)}, 2015.

\bibitem[Friedler et~al.(2016)Friedler, Scheidegger, and
  Venkatasubramanian]{friedler2016possibility}
S.~Friedler, C.~Scheidegger, and S.~Venkatasubramanian.
\newblock On the (im)possibility of fairness.
\newblock arXiv: 1609.07236 [cs.CY], 2016.

\bibitem[Garey and Johnson(1979)]{garey_comp_and_intractability}
M.~R. Garey and D.~S. Johnson.
\newblock \emph{Computers and Intractability: A Guide to the Theory of
  NP-Completeness}.
\newblock W. H. Freeman and Company, 1979.

\bibitem[Gillen et~al.(2018)Gillen, Jung, Kearns, and Roth]{gillen2018}
S.~Gillen, C.~Jung, M.~Kearns, and A.~Roth.
\newblock Online learning with an unknown fairness metric.
\newblock In \emph{Neural Information Processing Systems (NeurIPS)}, 2018.

\bibitem[Gonzalez(1985)]{gonzalez1985}
T.~F. Gonzalez.
\newblock Clustering to minimize the maximum intercluster distance.
\newblock \emph{Theoretical Computer Science}, 38:\penalty0 293--306, 1985.

\bibitem[Gottlob et~al.(2005)Gottlob, Greco, and Scarcello]{gottlob2005}
G.~Gottlob, G.~Greco, and F.~Scarcello.
\newblock Pure nash equilibria: Hard and easy games.
\newblock \emph{Journal of Artificial Intelligence Research}, 24:\penalty0
  357--406, 2005.

\bibitem[Hastie et~al.(2009)Hastie, Tibshirani, and Friedman]{hastie2009}
T.~Hastie, R.~Tibshirani, and J.~Friedman.
\newblock \emph{The Elements of Statistical Learning --- Data Mining,
  Inference, and Prediction}.
\newblock Springer, 2nd edition, 2009.

\bibitem[H\'{e}bert-Johnson et~al.(2018)H\'{e}bert-Johnson, Kim, Reingold, and
  Rothblum]{Hebert2018}
\'{U}. H\'{e}bert-Johnson, M.~P. Kim, O.~Reingold, and G.~N. Rothblum.
\newblock Calibration for the (computationally-identifiable) masses.
\newblock In \emph{International Conference on Machine Learning (ICML)}, 2018.

\bibitem[Huang et~al.(2019)Huang, Jiang, and Vishnoi]{huang2019}
L.~Huang, S.~H.-C. Jiang, and N.~K. Vishnoi.
\newblock Coresets for clustering with fairness constraints.
\newblock In \emph{Neural Information Processing Systems (NeurIPS)}, 2019.

\bibitem[Joseph et~al.(2016)Joseph, Kearns, Morgenstern, and Roth]{joseph2016}
M.~Joseph, M.~Kearns, J.~Morgenstern, and A.~Roth.
\newblock Fairness in learning: Classic and contextual bandits.
\newblock In \emph{Neural Information Processing Systems (NIPS)}, 2016.

\bibitem[Joseph et~al.(2018)Joseph, Kearns, Morgenstern, Neel, and
  Roth]{joseph2018}
M.~Joseph, M.~Kearns, J.~Morgenstern, S.~Neel, and A.~Roth.
\newblock Meritocratic fairness for infinite and contextual bandits.
\newblock In \emph{AAAI / ACM Conference on Artificial Intelligence, Ethics,
  and Society}, 2018.

\bibitem[Jung et~al.(2020)Jung, Kannan, and Lutz]{jung2020}
C.~Jung, S.~Kannan, and N.~Lutz.
\newblock A center in your neighborhood: Fairness in facility location.
\newblock In \emph{Symposium on Foundations of Responsible Computing (FORC)},
  2020.

\bibitem[Kearns et~al.(2018)Kearns, Neel, and Roth]{kearns_2018_subgroup1}
M.~Kearns, S.~Neel, and Z.~S. Roth, A.~Wu.
\newblock Preventing fairness gerrymandering: Auditing and learning for
  subgroup fairness.
\newblock In \emph{International Conference on Machine Learning (ICML)}, 2018.

\bibitem[Kearns et~al.(2019{\natexlab{a}})Kearns, Neel, and
  Roth]{kearns_2018_subgroup2}
M.~Kearns, S.~Neel, and Z.~S. Roth, A.~Wu.
\newblock An empirical study of rich subgroup fairness for machine learning.
\newblock In \emph{Conference on Fairness, Accountability, and Transparency
  (ACM FAT*)}, 2019{\natexlab{a}}.

\bibitem[Kearns et~al.(2019{\natexlab{b}})Kearns, Roth, and
  Sharifi-Malvajerdi]{kearns_average_individual_fairness}
M.~Kearns, A.~Roth, and S.~Sharifi-Malvajerdi.
\newblock Average individual fairness: Algorithms, generalization and
  experiments.
\newblock In \emph{Neural Information Processing Systems (NeurIPS)},
  2019{\natexlab{b}}.

\bibitem[Kim et~al.(2019)Kim, Ghorbani, and Zou]{kim2019}
M.~P. Kim, A.~Ghorbani, and J.~Zou.
\newblock Multiaccuracy: Black-box post-processing for fairness in
  classification.
\newblock In \emph{AAAI / ACM Conference on Artificial Intelligence, Ethics,
  and Society}, 2019.

\bibitem[Kleindessner et~al.(2019{\natexlab{a}})Kleindessner, Awasthi, and
  Morgenstern]{fair_k_center_2019}
M.~Kleindessner, P.~Awasthi, and J.~Morgenstern.
\newblock Fair $k$-center clustering for data summarization.
\newblock In \emph{International Conference on Machine Learning (ICML)},
  2019{\natexlab{a}}.
\newblock Code available on
  \url{https://github.com/matthklein/fair_k_center_clustering}.

\bibitem[Kleindessner et~al.(2019{\natexlab{b}})Kleindessner, Samadi, Awasthi,
  and Morgenstern]{fair_SC_2019}
M.~Kleindessner, S.~Samadi, P.~Awasthi, and J.~Morgenstern.
\newblock Guarantees for spectral clustering with fairness constraints.
\newblock In \emph{International Conference on Machine Learning (ICML)},
  2019{\natexlab{b}}.

\bibitem[Mahabadi and Vakilian(2020)]{mahabadi2020}
S.~Mahabadi and A.~Vakilian.
\newblock (individual) fairness for $k$-clustering.
\newblock arXiv:2002.06742 [cs.DS], 2020.

\bibitem[R\"{o}sner and Schmidt(2018)]{roesner2018}
C.~R\"{o}sner and M.~Schmidt.
\newblock Privacy preserving clustering with constraints.
\newblock In \emph{International Colloquium on Automata, Languages, and
  Programming (ICALP)}, 2018.

\bibitem[Schmidt et~al.(2018)Schmidt, Schwiegelshohn, and
  Sohler]{sohler_kmeans}
M.~Schmidt, C.~Schwiegelshohn, and C.~Sohler.
\newblock Fair coresets and streaming algorithms for fair k-means clustering.
\newblock arXiv:1812.10854 [cs.DS], 2018.

\bibitem[Shalev-Shwartz and Ben-David(2014)]{shalev2014understanding}
S.~Shalev-Shwartz and S.~Ben-David.
\newblock \emph{Understanding machine learning: From theory to algorithms}.
\newblock Cambridge University Press, 2014.

\bibitem[von Luxburg(2007)]{Luxburg_tutorial}
U.~von Luxburg.
\newblock A tutorial on spectral clustering.
\newblock \emph{Statistics and Computing}, 17\penalty0 (4):\penalty0 395--416,
  2007.

\bibitem[von Luxburg et~al.(2012)von Luxburg, Williamson, and
  Guyon]{luxburg_ScienceOrArt}
U.~von Luxburg, R.~Williamson, and I.~Guyon.
\newblock Clustering: Science or art?
\newblock In \emph{Workshop on Unsupervised and Transfer Learning}, 2012.

\bibitem[Wagstaff et~al.(2001)Wagstaff, Cardie, Rogers, and
  Schr{\"o}dl]{wagstaff2001constrained}
K.~Wagstaff, C.~Cardie, S.~Rogers, and S.~Schr{\"o}dl.
\newblock Constrained k-means clustering with background knowledge.
\newblock In \emph{International Conference on Machine Learning (ICML)}, 2001.

\end{thebibliography}
\bibliographystyle{plainnat}

\clearpage

\appendix

\section*{Appendix}\label{appendix}

\section{Related Work and Concepts}\label{appendix_related_work}

\paragraph{Existing Notions of Individual Fairness}

As 
discussed in Section~\ref{section_introduction}, the existing notions of fairness in 
ML, in particular in the context of classification, can largely be categorized into  group fairness and individual fairness.  
There is also a recent line of work on 
the notion of 
\emph{rich subgroup fairness} \citep{Hebert2018,kearns_2018_subgroup1,kearns_2018_subgroup2,kim2019}, 
which 
falls between these two categories
in that it requires some statistic to be similar 
for a \emph{large} (or even infinite) number of subgroups. Here we focus on the work strictly falling into the category of individual fairness. 

\citet{fta2012} were the first to provide a notion of individual fairness by asking that 
similar data points (as measured by a given task-specific metric) should be treated
similarly by a randomized classifier. 
\citet{joseph2016} and \citet{joseph2018} study fairness in multi-armed bandit problems. Their fairness notion  
 aims at guaranteeing fairness on the individual level by asking that in any round, an arm with a higher expected reward 
 (corresponding to a better qualified applicant, 
 for example) 
 is more likely to be played than an arm with a lower expected reward. 
Specifically in the contextual bandit setting, \citet{gillen2018} 
 apply the  principle of \citeauthor{fta2012} by requiring that in any round, 
 similar contexts are 
 picked  with approximately equal probability. 
The recent work of \citet{kearns_average_individual_fairness} studies the scenario that every individual is subject to a 
multitude 
of classification tasks 
and introduces the notion of average individual fairness. It asks that all individuals are classified with the same accuracy \emph{on average} over all classification~tasks.

\paragraph{Fairness for Clustering}

The most established notion of fairness for clustering has been proposed by 
\citet{fair_clustering_Nips2017}. 
It is based on the fairness notion of disparate impact
\citep{feldman2015}, which says that the output of a 
ML 
algorithm should be independent of a sensitive attribute, and 
asks that
each cluster has proportional representation from different
demographic groups.
\citeauthor{fair_clustering_Nips2017}
provide approximation algorithms that incorporate
their
notion 
into
$k$-center and $k$-median clustering, assuming that there are only two 
demographic 
groups. 
Several follow-up works extend this line of work to other 
clustering objectives such as $k$-means or spectral clustering, 
multiple or non-disjoint groups, 
some variations of the fairness notion  
or to address scalability issues \citep{roesner2018,sohler_kmeans,ahmadian2019,anagnostopoulos2019,backurs2019,bera2019,bercea2019,huang2019,fair_SC_2019}. 
The recent work of \citet{davidson2020} shows that for two groups, when given any clustering, one can efficiently compute the fair clustering 
(fair according to the notion of \citeauthor{fair_clustering_Nips2017})
that is most similar to the given clustering  
using 
linear programming.
\citeauthor{davidson2020} also show that it is NP-hard to decide whether a data set allows for a fair clustering that 
additionally 
satisfies some given must-link constraints.
They mention that such must-link constraints could be used for encoding individual level fairness constraints of the form ``similar data points must go to the same cluster''. 
However, for such a notion of individual fairness it remains unclear which pairs of data points exactly should be subject to a must-link constraint.

Three 
alternative fairness notions for clustering 
are
tied to centroid-based clustering such as $k$-means, $k$-center and $k$-median, 
 where one chooses $k$ centers and then forms clusters by assigning every data point to its closest center.
(i) Motivated 
by the 
application of 
data summarization, \citet{fair_k_center_2019} propose 
that the various demographic groups should be 
proportionally represented among 
the chosen centers.  
(ii) \citet{chen2019} propose a notion of proportionality that requires that 
no sufficiently large subset of data points could jointly reduce 
their distances from their 
closest 
centers by choosing a new center.   
The latter notion is similar to our notion of individual fairness in that it 
assumes that an individual data point 
strives 
to be well represented (in the notion of \citeauthor{chen2019} by being close to a center). Like our notion and other than the fairness notions of 
\citet{fair_clustering_Nips2017} and \citet{fair_k_center_2019}, it does not rely on demographic group information. 
However, while our notion aims at ensuring fairness for every single data point, the notion of \citeauthor{chen2019} only looks at sufficient large subsets. Furthermore, 
since our notion defines 
``being well represented'' in terms of the average distance of a data point to the other points in its cluster, 
our notion is not restricted to centroid-based clustering.
%
(iii) Only recently, 
\citet{jung2020}  proposed a notion of individual fairness for centroid-based clustering that comes with a guarantee for every single data point. It asks that every data point is 
somewhat close to a center, where ``somewhat'' depends on how close the data point is to its $k$ nearest neighbors. Building on the work of \citeauthor{jung2020}, \citet{mahabadi2020} proposed a local search based algorithm 
for this fairness notion that comes with constant factor approximation guarantees.

\paragraph{Average Attraction Property}

\citet{vempala2008} study which properties of a similarity function are sufficient in order to approximately recover 
(in either a list 
or a tree model) 
an unknown 
ground-truth clustering. One of the weaker properties they consider is 
the  average attraction property, 
which is 
closely related to 
our notion of individual fairness and requires inequality~\eqref{def_individual_fairness_ineq} to hold 
for the ground-truth clustering 
with 
an additive 
gap of~$\gamma>0$ between the 
left and the right side 
of 
\eqref{def_individual_fairness_ineq}. 
\citeauthor{vempala2008} show that the average attraction property is 
sufficient to successfully cluster in the list model, but with the length of the list 
being exponential in $1/\gamma$, 
and is not sufficient to successfully cluster in the tree model.
The conceptual difference between the work of \citeauthor{vempala2008} and 
ours 
is that the 
former assumes a ground-truth clustering and considers the 
average attraction property as a helpful property 
to find this ground-truth clustering, while we consider individual fairness as a constraint we would 
like to impose on whatever clustering~we~compute.

\paragraph{Game-theoretic Interpretation}

Fixing the number of clusters~$k$, 
our notion of an individually fair clustering can be interpreted in terms of a 
strategic game:   
let 
each data point 
correspond 
to a player that can play an action in $[k]$ in order to determine which cluster it belongs to. If, upon the cluster choice of each player,
a data point is treated fair according to Definition~\ref{def_indi_fairness}, this data point gets a utility value of $+1$; otherwise it gets a utility value of $0$. 
Then 
a clustering is individually fair if and only if it is a pure (strong / Pareto) Nash equilibrium of this 
particular 
game. It is well-known 
for many games 
that deciding whether the game has a pure Nash equilibrium is NP-hard \citep{gottlob2005}. However, none of the existing 
NP-hardness results in game theory 
implies NP-hardness of 
individually fair clustering.

\section{Proof of Theorem~\ref{theorem_hardness}}\label{proof_hardness}

We show NP-hardness 
of the individually fair clustering decision problem 
(with $k=2$ 
and $d$ required to be a metric) 
via a reduction from a variant of 3-SAT.   
It is well known that deciding whether a Boolean formula in conjunctive normal form, where each clause 
comprises at most three literals, is satisfiable is NP-hard. NP-hardness also holds for a restricted version of 3-SAT, where each variable 
occurs in at most three clauses \citep[][page 259]{garey_comp_and_intractability}. Furthermore, we can require the formula to have the 
same number of clauses as number of variables as the following transformation shows:
let $\Phi$ be a formula with $m$ clauses and $n$ variables. If $n>m$, we introduce $l=\lfloor\frac{n-m+1}{2}\rfloor$ new variables $x_1,\ldots,x_l$ 
and for each of them add three clauses $(x_i)$ to $\Phi$ (if $n-m$ is odd, we add only two clauses $(x_l)$). The resulting formula has the 
same number of clauses as number of variables and is satisfiable if and only if $\Phi$ is satisfiable. Similarly, if $n<m$, we introduce $l=\lfloor3\cdot\frac{m-n}{2}+\frac{1}{2}\rfloor$ 
new variables $x_1,\ldots,x_l$ and add to $\Phi$ the clauses $(x_1\vee x_2\vee x_3),(x_4\vee x_5\vee x_6),\ldots,(x_{l-2}\vee x_{l-1}\vee x_l)$ (if $m-n$ is odd, the last clause is 
$(x_{l-1}\vee x_l)$ instead of $(x_{l-2}\vee x_{l-1}\vee x_l)$). As before, the resulting formula has the 
same number of clauses as number of variables and is satisfiable if and only if $\Phi$ is satisfiable.

So let $\Phi=C_1 \wedge C_2\wedge \ldots \wedge C_n$ be a formula in conjunctive normal form over variables $x_1,\ldots,x_n$ 
such that each clause~$C_i$ comprises at most three literals 
$x_j$ or $\neg x_j$ and each variable occurs in at most three clauses (as either $x_j$ or $\neg x_j$). 
We construct a metric space $(\dataset,d)$ in time polynomial in $n$ such that $\dataset$ has an 
individually fair 
2-clustering with respect to $d$ if and only if $\Phi$ is satisfiable 
(for $n$ sufficiently large). 
We set 
\begin{align*}
\dataset=\{True, False,\star,\infty,C_1,\ldots,C_n,x_1,\neg x_1,\ldots,x_n,\neg x_n\} 
\end{align*}
and 
\begin{align*}
d(x,y)=\left[d'(x,y)+\indi\{x\neq y\}\right]+\indi\{x\neq y\}\cdot\max_{x,y\in\dataset}\left[d'(x,y)+1\right],\quad x,y\in\dataset,
\end{align*}
for some symmetric function $d':\dataset\times\dataset\rightarrow\R_{\geq 0}$ with $d'(x,x)=0$, $x\in\dataset$, that we specify in the next paragraph. 
It is straightforward to see that $d$ is a metric. Importantly, note that for any $x\in\dataset$, inequality \eqref{def_individual_fairness_ineq} holds with respect to $d$ if and only 
if it holds with respect to $d'$.

We set $d'(x,y)=0$ for all $x,y\in\dataset$ except for the following:
\begin{align*}
d'(True,False)&=A,\\
d'(True,\star)&=B,\\
d'(\star,False)&=C,\\
d'(C_i,False)&=D,\quad i=1,\ldots,n,\\
d'(C_i,\star)&=E,\quad i=1,\ldots,n,\\
d'(\infty,True)&=F,\\
d'(\infty,False)&=G,\\
d'(\infty,\star)&=H,\\
d'(C_i,\infty)&=J,\quad i=1,\ldots,n,\\
d'(x_i,\neg x_i)&=S,\quad i=1,\ldots,n,\\
d'(C_i,\neg x_j)&=U, \quad  (i,j)\in\{(i,j)\in\{1,\ldots,n\}^2:x_j\text{~appears in~}C_i\},\\
d'(C_i, x_j)&=U, \quad (i,j)\in\{(i,j)\in\{1,\ldots,n\}^2:\neg x_j\text{~appears in~}C_i\},
\end{align*}
where we set
\begin{align}\label{choice_numbers}
\begin{split}
 A &=  n, \qquad F = n^2, \qquad B = 2F=2n^2, \qquad E = \frac{5}{2}F = \frac{5}{2} n^2, \\
 J &= E+\log n=\frac{5}{2} n^2+\log n, \qquad D = J + \log^2 n=\frac{5}{2} n^2+\log n+\log^2 n,\\
 U &=3J=\frac{15}{2} n^2+3\log n, \qquad H= n D +E=\frac{5}{2} n^3+\frac{5}{2} n^2+n\log n+n\log^2 n,\\
G &= H+2n^2-n-2n\log^2 n=\frac{5}{2} n^3+\frac{9}{2} n^2-n+n\log n-n\log^2 n,\\
S &= (3n+3)U=\frac{45}{2} n^3+\frac{45}{2} n^2+9n\log n+9\log n,\\ 
C &= \frac{A + G + n  D}{2}=\frac{5}{2} n^3+\frac{9}{4} n^2+n\log n.
\end{split}
\end{align}

We 
show that 
for $n\geq 160$ 
there 
is 
a satisfying assignment 
for $\Phi$
if and only if there 
is 
an individually fair 2-clustering of  $\dataset$.

\begin{itemize}[leftmargin=*]
 \item  \emph{``Satisfying assignment $\Rightarrow$ individually fair 2-clustering''}

Let us assume we are given a satisfying assignment of $\Phi$. We may assume that if $x_i$ only appears as $x_i$ in $\Phi$ and not as $\neg x_i$, then $x_i$ is true; 
similarly, if $x_i$ only appears as $\neg x_i$, then $x_i$ is false.
We construct a clustering of $\dataset$ into two clusters $V_1$ and $V_2$ as follows:
\begin{align*}
 V_1&=\{True,\infty,C_1,\ldots,C_n\}\cup\{x_i: x_i\text{~is true in sat. ass.}\}\cup\{\neg x_i: \neg x_i\text{~is true in sat. ass.}\},\\
 V_2&=\{False,\star\}\cup\{x_i: x_i\text{~is false in satisfying assignment}\}\cup\{\neg x_i: \neg x_i\text{~is false in sat. ass.}\}.
\end{align*}
It is $|V_1|=2+2n$ and $|V_2|=2+n$. We need show that every data point in $\dataset$ is treated individually fair. This is equivalent to verifying that the following inequalities are true:

\vspace{2mm}
Points in $V_1$:
\begin{align}\label{cond_points_X1}
 True:~~~~&\frac{1}{1+2n}\sum_{v\in V_1}d'(True,v)=\frac{F}{1+2n}\leq\frac{A+B}{2+n}=\frac{1}{2+n}\sum_{v\in V_2}d'(True,v)\\
\infty:~~~~&\frac{1}{1+2n}\sum_{v\in V_1}d'(\infty,v)=\frac{F+nJ}{1+2n}\leq \frac{G+H}{2+n}= \frac{1}{2+n}\sum_{v\in V_2}d'(\infty,v)\\
C_i:~~~~& \frac{1}{1+2n}\sum_{v\in V_1}d'(C_i,v)\leq\frac{J+2U}{1+2n}\leq \frac{U+D+E}{2+n}\leq \frac{1}{2+n}\sum_{v\in V_2}d'(C_i,v)\\
x_i:~~~~&\frac{1}{1+2n}\sum_{v\in V_1}d'(x_i,v)\leq\frac{2U}{1+2n}\leq \frac{S}{2+n}\leq \frac{1}{2+n}\sum_{v\in V_2}d'(x_i,v)\\
\neg  x_i:~~~~&\frac{1}{1+2n}\sum_{v\in V_1}d'(\neg x_i,v)\leq\frac{2U}{1+2n}\leq \frac{S}{2+n}\leq \frac{1}{2+n}\sum_{v\in V_2}d'(\neg x_i,v)\label{cond_points_X1_end}
\end{align}

\vspace{2mm}
Points in $V_2$:
\begin{align}
False:~~~~&\frac{1}{1+n}\sum_{v\in V_2}d'(False,v)=\frac{C}{1+n}\leq\frac{A+G+nD}{2+2n}=\frac{1}{2+2n}\sum_{v\in V_1}d'(False,v)\label{cond_False_is_fair}\\
\star:~~~~&\frac{1}{1+n}\sum_{v\in V_2}d'(\star,v)=\frac{C}{1+n}\leq\frac{B+H+nE}{2+2n}=\frac{1}{2+2n}\sum_{v\in V_1}d'(\star,v)\\
x_i:~~~~&\frac{1}{1+n}\sum_{v\in V_2}d'(x_i,v)=0\leq \frac{S}{2+2n}\leq \frac{1}{2+2n}\sum_{v\in V_1}d'(x_i,v)\\
\neg  x_i:~~~~&\frac{1}{1+n}\sum_{v\in V_2}d'(\neg x_i,v)=0\leq \frac{S}{2+2n}\leq \frac{1}{2+2n}\sum_{v\in V_1}d'(\neg x_i,v)\label{cond_points_X2_end}
\end{align}

It is straightforward to check that for our choice of $A,B,C,D,E,F,G,H,J,S,U$ as specified 
in \eqref{choice_numbers} all inequalities~\eqref{cond_points_X1} to \eqref{cond_points_X2_end} are true.

\vspace{2mm}
\item \emph{``Individually fair 2-clustering $\Rightarrow$ satisfying assignment''}

Let us assume that there is an individually fair clustering of $\dataset$ with two clusters $V_1$ and $V_2$.
For any partitioning of $\{C_1,\ldots,C_n\}$ into two sets of size $l$ and $n-l$ ($0\leq l\leq n$) we denote the two sets by $\mathcal{C}_l$ and $\widetilde{\mathcal{C}}_{n-l}$.

We first show that  $x_i$ and $\neg x_i$ cannot be contained in the same cluster (say in $V_1$). This is because if we assume that $x_i,\neg x_i \in V_1$, 
for our choice of $S$ and $U$ in \eqref{choice_numbers} we have
\begin{align*}
\frac{1}{|V_2|}\sum_{v\in V_2}d'(x_i,v)\leq U< \frac{S}{3n+2}\leq \frac{1}{|V_1|-1}\sum_{v\in V_1}d'(x_i,v)
\end{align*}
in contradiction to $x_i$ being treated individually fair. As a consequence we have $n\leq |V_1|,|V_2|\leq 2n+4$.

Next, we show that due to our choice of $A,B,C,D,E,F,G,H,J$ in \eqref{choice_numbers} none of the following cases can be true:

\vspace{1mm}
\begin{enumerate}
\setlength\itemsep{5mm}

 \item $\{True,\infty\}\cup \mathcal{C}_l\subset V_1$ and $\widetilde{\mathcal{C}}_{n-l}\cup\{False,\star\}\subset V_2$ 
 for any $0\leq l< n$

 \vspace{2.4mm}
In this case, $False$ would not be treated fair since for all $0\leq l< n$,
\begin{align*}
\frac{1}{|V_1|}\sum_{v\in V_1}d'(False,v)= \frac{A+G+lD}{l+2+n}<\frac{C+(n-l)D}{n-l+1+n}=\frac{1}{|V_2|-1}\sum_{v\in V_2}d'(False,v).
\end{align*}

 \item $\{True\}\cup \mathcal{C}_l\subset V_1$ and $\widetilde{\mathcal{C}}_{n-l}\cup\{False,\star,\infty\}\subset V_2$ for any $0\leq l\leq n$

 \vspace{2.4mm}
 In this case, $False$ would not be treated fair since for all $0\leq l\leq n$,
\begin{align*}
 \frac{1}{|V_1|}\sum_{v\in V_1}d'(False,v)=\frac{A+lD}{l+1+n}<\frac{C+G+(n-l)D}{n-l+2+n}=\frac{1}{|V_2|-1}\sum_{v\in V_2}d'(False,v). 
\end{align*}

\item $\{False,\infty\}\cup\mathcal{C}_l\subset V_1$ and $\widetilde{\mathcal{C}}_{n-l}\cup\{True,\star\}\subset V_2$ for any $0\leq l\leq n$

\vspace{2.4mm}
In this case, $True$ would not be treated fair since for all $0\leq l\leq  n$,
 \begin{align*}
\frac{1}{|V_1|}\sum_{v\in V_1}d'(True,v)= \frac{A+F}{l+2+n}<\frac{B}{n-l+1+n}=\frac{1}{|V_2|-1}\sum_{v\in V_2}d'(True,v).
\end{align*}

\item $\{False\}\cup\mathcal{C}_l\subset V_1$ and $\widetilde{\mathcal{C}}_{n-l}\cup\{True,\star,\infty\}\subset V_2$ for any $0\leq l\leq n$

\vspace{2.4mm}
In this case, $True$ would not be treated fair since for all $0\leq l\leq  n$, 
\begin{align*}
 \frac{1}{|V_1|}\sum_{v\in V_1}d'(True,v)=\frac{A}{l+1+n}<\frac{B+F}{n-l+2+n}=\frac{1}{|V_2|-1}\sum_{v\in V_2}d'(True,v).
\end{align*}

\item $\{\star,\infty\}\cup\mathcal{C}_l\subset V_1$ and $\widetilde{\mathcal{C}}_{n-l}\cup\{False,True\}\subset V_2$ for any $0\leq l\leq n$

\vspace{2.4mm}
In this case, $\star$ would not be treated fair since for all $0\leq l\leq  n$,
\begin{align*}
\frac{1}{|V_2|}\sum_{v\in V_2}d'(\star,v)= \frac{B+C+(n-l)E}{n-l+2+n}<\frac{H+lE}{l+1+n}=\frac{1}{|V_1|-1}\sum_{v\in V_1}d'(\star,v). 
\end{align*}

\item $\{\star\}\cup\mathcal{C}_l\subset V_1$ and $\widetilde{\mathcal{C}}_{n-l}\cup\{False,True,\infty\}\subset V_2$ for any $0\leq l\leq n$

\vspace{2.4mm}
In this case, $\infty$ would not be treated fair since for all $0\leq l\leq n$,
\begin{align*}
\frac{1}{|V_1|}\sum_{v\in V_1}d'(\infty,v)=\frac{H+lJ}{l+1+n}<\frac{F+G+(n-l)J}{n-l+2+n}=\frac{1}{|V_2|-1}\sum_{v\in V_2}d'(\infty,v).
\end{align*}

\item $\mathcal{C}_l\subseteq V_1$ and $\widetilde{\mathcal{C}}_{n-l}\cup\{True,False,\star,\infty\}\subseteq V_2$ for any $0\leq l\leq n$
 
 \vspace{2.4mm}
 In this case, $True$ would not be treated fair since for all $0\leq l\leq  n$,
\begin{align*}
  \frac{1}{|V_1|}\sum_{v\in V_1}d'(True,v)=0<\frac{A+B+F}{3+(n-l)+n}=\frac{1}{|V_2|-1}\sum_{v\in V_2}d'(True,v).  
\end{align*}

\item $\{\infty\}\cup\mathcal{C}_l\subseteq V_1$ and $\widetilde{\mathcal{C}}_{n-l}\cup\{True,False,\star\}\subseteq V_2$ for any $0\leq l\leq n$

\vspace{2.4mm}
In this case, $True$ would not be treated fair since for all $0\leq l\leq  n$,
\begin{align*}
  \frac{1}{|V_1|}\sum_{v\in V_1}d'(True,v)=\frac{F}{1+l+n}<\frac{A+B}{2+(n-l)+n}=\frac{1}{|V_2|-1}\sum_{v\in V_2}d'(True,v).
\end{align*}

\end{enumerate}

\vspace{2mm}
Of course, in all these cases we can exchange the role of $V_1$ and $V_2$. Hence, $True,\infty, C_1,\ldots,C_n$ must be contained in one cluster and $\star,False$ must be 
contained in the other cluster. W.l.o.g., let us assume $True,\infty, C_1,\ldots,C_n\in V_1$ and  $\star,False\in V_2$ and hence 
$|V_1|=2n+2$ and $|V_2|=n+2$.

Finally, we show that for the clause $C_i=(l_j)$ or $C_i=(l_j\vee l_{j'})$ or $C_i=(l_j\vee l_{j'} \vee l_{j''})$, with the literal $l_j$ equaling $x_j$ or $\neg x_j$,  
it cannot be the case that $C_i,\neg l_j$ or  $C_i,\neg l_j$, $\neg l_{j'}$ or  $C_i,\neg l_j$, $\neg l_{j'}$, $\neg l_{j''}$ are all contained in $V_1$. 
This is because otherwise
\begin{align}\label{wider8}
 \frac{1}{|V_2|}\sum_{v\in V_2}d'(C_i,v)=\frac{D+E}{n+2}<\frac{U+J}{2n+1}\leq\frac{1}{|V_1|-1}\sum_{v\in V_1}d'(C_i,v)
\end{align}
for our choice of $D,E,J,U$ in \eqref{choice_numbers} and $C_i$ would not be treated fair. Consequently, since $x_j$ and $\neg x_j$ are not in the same cluster, 
for each clause $C_i$ at least one of its literals must be in~$V_1$. 

Hence, if we set every literal $x_i$ or $\neg x_i$ that is contained in $V_1$ to a true logical value 
and every literal $x_i$ or $\neg x_i$ that is contained in~$V_2$ to a false logical value, we obtain a valid assignment 
that makes $\Phi$ true.  
\hfill $\square$
\end{itemize}

\vspace{6mm}
\section{Proof of Lemma~\ref{lemma_boundary_points} and Theorem~\ref{lemma_existence_1D}}\label{proof_existence}

We assume that $\dataset=\{x_1,\ldots,x_n\}\subseteq \R$ with $x_1\leq x_2\leq \ldots \leq x_n$ and write the Euclidean metric $d(x_i,x_j)$ between two points $x_i$ and $x_j$  
in its usual way $|x_i-x_j|$. We first prove Lemma~\ref{lemma_boundary_points}.

\vspace{2mm}
\textbf{Proof of Lemma~\ref{lemma_boundary_points}:}

If $\mathcal{C}$ is  fair, then all points $x_{i_l}$ and $x_{i_l+1}$, $l\in[k-1]$, are treated fair. 
Conversely, let us assume that $x_{i_l}$ and $x_{i_l+1}$, $l\in[k-1]$, are treated fair. We need to show that all points in $\dataset$ are treated fair.  
Let $\tilde{x}\in C_l=\{x_{i_{l-1}+1},\ldots,x_{i_l}\}$ for some $l\in\{2,\ldots,k-1\}$ 
and $l'\in\{l+1,\ldots,k\}$. Since $x_{i_l}$ is treated 
fair, we have
\begin{align*}
 \frac{1}{|C_l|-1}\sum_{y\in{C_{l}}} (x_{i_l}-y)= \frac{1}{|C_l|-1}\sum_{y\in{C_{l}}} |x_{i_l}-y|\leq  
  \frac{1}{|C_{l'}|}\sum_{y\in{C_{l'}}} |x_{i_l}-y|=\frac{1}{|C_{l'}|}\sum_{y\in{C_{l'}}} (y-x_{i_l}) 
\end{align*}
and hence
\begin{align*}
 \frac{1}{|C_l|-1}\sum_{y\in{C_{l}}} |\tilde{x}-y|&\leq  \frac{1}{|C_l|-1}\sum_{y\in C_{l}\setminus\{\tilde{x}\}} (|\tilde{x}-x_{i_l}|+|x_{i_l}-y|)\\
 &= (x_{i_l}-\tilde{x})+\frac{1}{|C_l|-1}\sum_{y\in C_{l}\setminus\{\tilde{x}\}} (x_{i_l}-y)\\
 &\leq (x_{i_l}-\tilde{x})+\frac{1}{|C_{l'}|}\sum_{y\in{C_{l'}}} (y-x_{i_l})\\
 &=\frac{1}{|C_{l'}|}\sum_{y\in{C_{l'}}} (y-\tilde{x})\\
 &=\frac{1}{|C_{l'}|}\sum_{y\in{C_{l'}}} |\tilde{x}-y|. 
\end{align*}
Similarly, we can show for $l'\in\{1,\ldots,l-1\}$ that 
\begin{align*}
 \frac{1}{|C_l|-1}\sum_{y\in{C_{l}}} |\tilde{x}-y|\leq  \frac{1}{|C_{l'}|}\sum_{y\in{C_{l'}}} |\tilde{x}-y|, 
\end{align*}
and hence $\tilde{x}$ is treated 
fair. 
Similarly, we can show that all points $x_1,\ldots,x_{{i_1}-1}$ and $x_{i_{k-1}+2},\ldots,x_n$ are treated 
fair.

For the second claim observe that for $1\leq s\leq l-1$, the average distance of $x_{i_l}$ to the points in $C_s$ cannot 
 be smaller than the average distance to the points 
in $C_l\setminus\{x_{i_l}\}$
and for $l+2\leq s\leq k$,  
the average distance of $x_{i_l}$ to the points in $C_s$ cannot be smaller than the average distance to the points 
in $C_{l+1}$. A similar argument proves the claim for $x_{i_l+1}$. \hfill$\square$

\vspace{6mm}
For $k=1$, $\mathcal{C}=(\dataset)$ is an individually fair $k$-clustering of $\dataset$ with contiguous clusters, and Theorem~\ref{lemma_existence_1D} is vacuously true.  In order to prove 
Theorem~\ref{lemma_existence_1D} for $k\geq 2$, we present an algorithm to compute an individually fair $k$-clustering of $\dataset$ with contiguous clusters. Our algorithm maintains an array $T$ 
of $k-1$ strictly 
increasing boundary indices that specify the right-most points of the first $k-1$ clusters. Starting from $T=(1,2,\ldots,k-1)$, corresponding to the 
clustering $(\{x_1\},\{x_2\},\ldots,\{x_{k-1}\},\{x_k,x_{k+1},\ldots,x_n\})$, it keeps incrementing the entries of $T$ until a fair clustering has been found. 
We formally state our algorithm as Algorithm~\ref{alg_1D} below.

In order to prove Theorem~\ref{lemma_existence_1D}, 
we 
need to 
show that Algorithm~\ref{alg_1D} always terminates and outputs an increasingly sorted array 
$T=(T[1],\ldots,T[k-1])$ with $1\leq T[1]<T[2]<\ldots<T[k-1]<n$ that  defines an individually fair clustering 
(obviously, the output $T$ defines a $k$-clustering with contiguous clusters). 
For doing so, 
we show several claims to be true.

\vspace{3mm}
\emph{Claim 1: Throughout the execution of Algorithm~\ref{alg_1D} we have $T[j]<T[j+1]$ for all 
$j\in[k-2]$.
}

\vspace{-1mm}
This is true at the beginning of the execution. Assume it is true before an update of $T$ happens. If $T[k-1]$ is updated, it is still true after the update. 
If $T[j_0]$ for some $j_0\in[k-2]$ is updated, we have $0\leq AvgDist_{Not}(x_{T[j_0]+1},C_{j_0}^T)<AvgDist_{In}(x_{T[j_0]+1},C_{j_0+1}^T)$ before the update. But then 
it is $C_{j_0+1}^T\supsetneq \{x_{T[j_0]+1}\}$ and $T[j_0+1]>T[j_0]+1$ before the update. Hence, also after the update of $T[j_0]$ the claim is true.

\vspace{3mm}
\emph{Claim 2: Throughout the execution of Algorithm~\ref{alg_1D} we have $T[k-1]\leq n-1$.}

\vspace{-1mm}
Assume that $T[k-1]=n-1$ would be updated to $T[k-1]=n$. But then, before the update, $C_k^T=\{x_n\}$ and $0\leq AvgDist_{Not}(x_{n},C_{k-1}^T)<AvgDist_{In}(x_{n},C_{k}^T)$.  However, 
$AvgDist_{In}(x_{n},\{x_n\})=0$.

\vspace{3mm}
From Claim~1 and Claim~2 it follows that Algorithm~\ref{alg_1D} terminates after at most $\binom{n-1}{k-1}$ updates.

\vspace{3mm}
\emph{Claim 3:
For $j\in[k-1]$, after any update $T[j]=T[j]+1$ until the next update of $T[j]$, 
the point $x_{T[j]}$ (referring to the value of $T[j]$ after the update) is treated individually fair.} 

\vspace{-1mm}
Since $x_{T[j]+1}$ (referring to the value of $T[j]$ before the update; after the update this point becomes $x_{T[j]}$) is the left-most point in its cluster, 
the closest cluster for $x_{T[j]+1}$ is either its own cluster or the cluster left of its own cluster. If $T[j]$ is 
updated to $T[j]+1$, this just means that $x_{T[j]+1}$ is closer to the left cluster and is now assigned to this cluster. 
So immediately after the update,  $x_{T[j]}$ (referring to the value of $T[j]$ after the update) is treated individually fair. 
As long as $T[j]$ is not updated for another time, $x_{T[j]}$ is the right-most point in its cluster~$C^T_j$ and cannot be closer to any cluster $C^T_l$, 
$l\in[j-1]$, than to its own cluster, no matter how often $T[l]$, $l\in[j-1]$, is updated. If $T[l]$ for $l\in\{j+1,\ldots,k-1\}$ gets updated, then 
$AvgDist_{Not}(x_{T[j]},C^T_l)$ can get only larger, so that $x_{T[j]}$  is still treated individually fair.

\vspace{3mm}
\emph{Claim 4:
After the last update of $T$ in the execution of Algorithm~\ref{alg_1D} all points $x_{T[j]+1}$, $j\in\{1,\ldots,k-1\}$ are treated individually fair.} 

\vspace{-1mm}
After the last update, Algorithm~\ref{alg_1D} checks for every point $x_{T[j]+1}$, $j\in[k-1]$, whether it is closer to its own cluster 
or the cluster on its left side and confirms 
that it is closer to its own cluster. Since $x_{T[j]+1}$ is the left-most point in its cluster, this implies that  $x_{T[j]+1}$ is treated individually fair.

\vspace{3mm}
From Claim~3, Claim~4 and Lemma~\ref{lemma_boundary_points} it follows that the output of Algorithm~\ref{alg_1D} is an individually fair clustering.

\newpage

~
\vspace{0cm}
\begin{algorithm}[H]
   \caption{Algorithm for finding an individually fair clustering in the 1-dim Euclidean case}
   \label{alg_1D}
   
\begin{small}   
\begin{algorithmic}[1]
\vspace{1mm}
    \STATE   {\bfseries Input:} 
      increasingly sorted array $(x_1,\ldots,x_n)$ of $n$ distinct points in $\R$; number of clusters~$k\in\{2,\ldots,|\dataset|\}$

\vspace{1mm}  
   \STATE {\bfseries Output:} 
   increasingly sorted array $T=(T[1],\ldots,T[k-1])$ of $k-1$ distinct boundary indices $T[i]\in\{1,\ldots,n-1\}$ defining $k$ clusters as follows: 
   $C_1=\{x_1,\ldots,x_{T[1]}\}$, $C_2=\{x_{T[1]+1},\ldots,x_{T[2]}\}$, \ldots, $C_k=\{x_{T[k-1]+1},\ldots,x_{n}\}$ 
   
\vspace{3mm}
\STATE{\emph{\# Conventions:} 
\vspace{0mm}
\begin{itemize}[leftmargin=*]
\setlength{\itemsep}{1pt}
 \item \emph{for an array of boundary indices~$T$ as in Line~2, $(C_1^T,\ldots,C_k^T)$ denotes the clustering with clusters $C_i^T$ defined as in Line~2}
\item 
\emph{
for a cluster~$C_i^T$ and a point $y\notin C_i^T$, we write 
$$AvgDist_{Not}(y,C_i^T)=\frac{1}{|C_i^T|}\sum_{z\in C_i^T}|y-z|$$
}
\item 
\emph{
for a cluster~$C_i^T$ and a point $y\in C_i^T$, we write (using the convention that $\frac{0}{0}=0$)
$$AvgDist_{In}(y,C_i^T)=\frac{1}{|C_i^T|-1}\sum_{z\in C_i^T}|y-z|$$ 
}
\end{itemize}
}

\vspace{4mm}
\STATE{Initialize~~$T=(1,2,\ldots,k-1)$}
\vspace{2pt}
\STATE{Set $DistLeft=AvgDist_{Not}(x_{T[k-1]+1},C_{k-1}^T)$,~~~$DistOwn=AvgDist_{In}(x_{T[k-1]+1},C_{k}^T)$~~and\\ 
~~~$IsFairOuter=\indi\{DistOwn\leq DistLeft\}$}

\vspace{4pt}
\WHILE{$IsFairOuter==False$} 
\vspace{2pt}
\STATE{Update $T[k-1]=T[k-1]+1$} 
\vspace{2pt}
\STATE{Set $SomethingChanged=True$} 

\vspace{4pt}
\WHILE{$SomethingChanged==True$} 
\vspace{2pt}
\STATE{Set $SomethingChanged=False$}

\vspace{4pt}
\FOR{$j=k-2$ {\bfseries to} $j=1$ \textbf{by} $-1$} 
\vspace{2pt}
\STATE{Set $DistLeft=AvgDist_{Not}(x_{T[j]+1},C_{j}^T)$,~~~$DistOwn=AvgDist_{In}(x_{T[j]+1},C_{j+1}^T)$~~and\\
~~~$IsFairInner=\indi\{DistOwn\leq DistLeft\}$}

\vspace{4pt}
\WHILE{$IsFairInner==False$} 
\vspace{2pt}
\STATE{Update $T[j]=T[j]+1$} 
\vspace{2pt}
\STATE{Set $SomethingChanged=True$} 
\vspace{2pt}
\STATE{Set $DistLeft=AvgDist_{Not}(x_{T[j]+1},C_{j}^T)$,~~~$DistOwn=AvgDist_{In}(x_{T[j]+1},C_{j+1}^T)$~and\\
~~~$IsFairInner=\indi\{DistOwn\leq DistLeft\}$
}
\ENDWHILE
\vspace{4pt}
\ENDFOR
\vspace{4pt}
\ENDWHILE
 \vspace{4pt}
\STATE{Set $DistLeft=AvgDist_{Not}(x_{T[k-1]+1},C_{k-1}^T)$,~~~$DistOwn=AvgDist_{In}(x_{T[k-1]+1},C_{k}^T)$ and\\
~~~$IsFairOuter=\indi\{DistOwn\leq DistLeft\}$
}

\vspace{4pt}
\ENDWHILE
\vspace{4pt}
\RETURN~~$T$
\end{algorithmic}
\end{small}
\end{algorithm}


\newpage

\section{Explanation of the Recurrence Relation~\eqref{recurrence_relation} and 
Modifications of the Dynamic Programming Approach of Section~\ref{section_1dim} to the Case $p=\infty$}\label{appendix_p_equals_infty}

Let us first explain the recurrence relation~\eqref{recurrence_relation}:  
because of $\|(x_1,\ldots,x_l)\|_p^p=\|(x_1,\ldots,x_{l-1})\|_p^p+|x_l|^p$ and for every clustering~$(C_1,\ldots,C_l)\in\mathcal{H}_{i,j,l}$ it 
is $|C_l|=j$, we have
\begin{align}\label{recurrence_relation_explained}
T(i,j,l)=|j-t_l|^p+\min_{(C_1,\ldots,C_l)\in\mathcal{H}_{i,j,l}} \|(|C_1|-t_1,\ldots,|C_{l-1}|-t_{l-1})\|_p^p.
\end{align}
It follows from Lemma~\ref{lemma_boundary_points} that a clustering $(C_1,\ldots,C_l)$ of $\{x_1,\ldots,x_i\}$ with contiguous clusters and $C_l=\{x_{i-j+1},\ldots,x_i\}$ is 
fair if and only if $(C_1,\ldots,C_{l-1})$ is a fair clustering of  $\{x_1,\ldots,x_{i-j}\}$ and the average distance of $x_{i-j}$ to the points in 
$C_{l-1}\setminus\{x_{i-j}\}$ is not greater 
than  the average distance to the points in $C_l$ and the average distance of $x_{i-j+1}$ to the points in $C_{l}\setminus\{x_{i-j+1}\}$ is not greater 
than  the average distance to the points in $C_{l-1}$. The latter two conditions correspond to the two inequalities in~\eqref{recurrence_relation} 
(when 
$|C_{l-1}|=s$, where $s$ is a variable). 
By 
explicitly enforcing 
these 
two constraints,  
we can utilize the first condition and 
rather than minimizing over  $\mathcal{H}_{i,j,l}$ in \eqref{recurrence_relation_explained}, 
we can minimize over both $s\in[i-j-(l-2)]$ and $\mathcal{H}_{i-j,s,l-1}$ 
(corresponding to minimizing over all fair 
$(l-1)$-clusterings of  $\{x_1,\ldots,x_{i-j}\}$ with non-empty contiguous clusters). It is 
\begin{align*}
\min_{\substack{s\in[i-j-(l-2)]\\ (C_1,\ldots,C_{l-1})\in\mathcal{H}_{i-j,s,l-1}}} 
\|(|C_1|-t_1,\ldots,|C_{l-1}|-t_{l-1})\|_p^p =\min_{s\in[i-j-(l-2)]} T(i-j,s,l-1),
\end{align*}
and hence we end up with the recurrence relation~\eqref{recurrence_relation}.

\vspace{6mm}
Now we describe how to modify the dynamic programming approach of Section~\ref{section_1dim} to the case $p=\infty$: 
in this case, we replace the definition of the table~$T$ in \eqref{definition_table_T} by  
\begin{align*}
T(i,j,l)=\min_{(C_1,\ldots,C_l)\in\mathcal{H}_{i,j,l}} \|(|C_1|-t_1,\ldots,|C_l|-t_l)\|_{\infty},\quad i\in[n], j\in[n], l\in[k],
\end{align*}
and $T(i,j,l)=\infty$ if $\mathcal{H}_{i,j,l}=\emptyset$ as before.
The optimal value of \eqref{1dim-problem} is now given 
by $\min_{j\in[n]} T(n,j,k)$. Instead of \eqref{table_T_initial}, 
we have, for $i,j\in[n]$, 
\begin{align*}
T(i,j,1)=\begin{cases}
 |i-t_1|,& j=i, \\
 \infty, &j\neq i\\           
           \end{cases},\hspace{7mm}
T(i,j,i)=\begin{cases}
 \max_{s=1,\ldots,i}|1-t_s|,& j=1, \\
 \infty, &j\neq 1\\           
           \end{cases}
\end{align*}
and
\begin{align*}
T(i,j,l)=\infty,\quad j+l-1>i,
\end{align*}
and
 the recurrence relation~\eqref{recurrence_relation} now becomes, for $l>1$ and $j+l-1\leq i$,
\begin{align*}
&T(i,j,l)=\max\Bigg\{|j-t_l|,\min\bigg\{T(i-j,s,l-1):  s\in[i-j-(l-2)],\\
&~~~~~~~~~~~~~~~~~~~~~~~~~~~~~~~~~~~~~~~~~~~~~~~~~~~~~~\frac{1}{s-1}\,{\sum_{f=1}^{s-1}|x_{i-j}-x_{i-j-f}|}\leq \frac{1}{j}\,{\sum_{f=1}^{j}|x_{i-j}-x_{i-j+f}|},\\
&~~~~~~~~~~~~~~~~~~~~~~~~~~~~~~~~~~~~~~~~~~~~~~~~~~~~~~\frac{1}{j-1}\,{\sum_{f=2}^{j}|x_{i-j+1}-x_{i-j+f}|}\leq\frac{1}{s}\,{\sum_{f=0}^{s-1}|x_{i-j+1}-x_{i-j-f}|}\bigg\}\Bigg\}.
\end{align*}

Just like before, we can build the table~$T$ in 
time $\mathcal{O}(n^3k)$. Computing a solution $(C_1^*,\ldots,C_k^*)$ to \eqref{1dim-problem} 
also works similarly as before. The only thing that we have to change is the condition (i) on $h_0$ 
(when setting $|C_l^*|=h_0$ for $l=k-1,\ldots,2$): now $h_0$ must satisfy 
\begin{align*}
\max\left\{T\left(n-\sum_{r=l+1}^k |C_r^*|,h_0,l\right),\max_{r=l+1,\ldots,k} ||C_r^*|-t_r|\right\}=v^*
\end{align*}
or equivalently 
 \begin{align*}
 T\left(n-\sum_{r=l+1}^k |C_r^*|,h_0,l\right)\leq v^*.
 \end{align*}

\begin{figure}[t]
 \centering
 \includegraphics[scale=0.38]{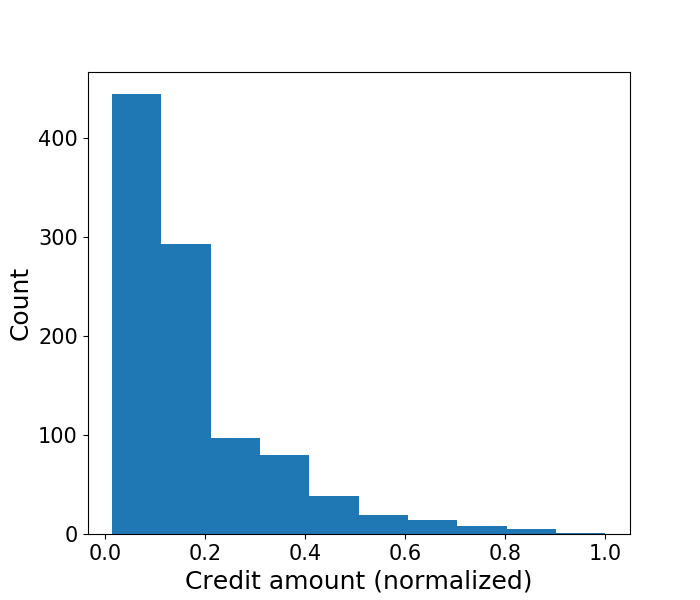}
 \hspace{0.2cm}
 \includegraphics[scale=0.38]{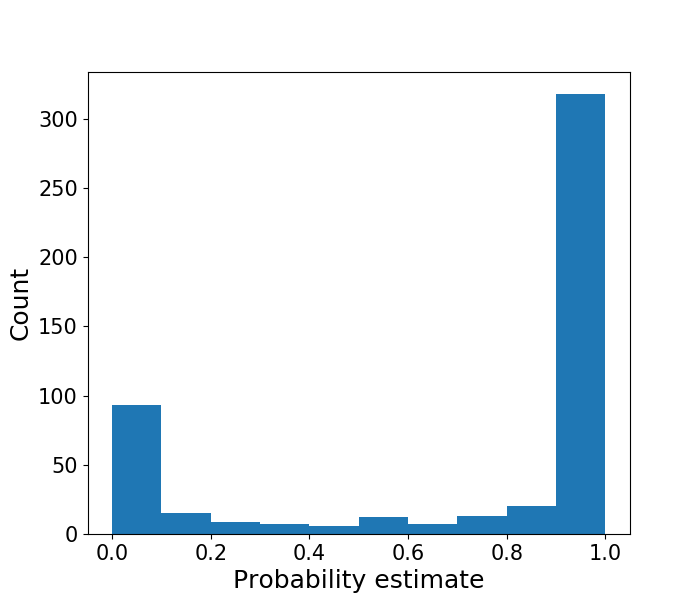}
 
 \caption{Histograms of the data sets used in the experiments of Section~\ref{section_experiments_1D}. \textbf{Left:} The credit amount 
 (one of the 20 features in the German credit data set; normalized to be in $[0,1]$) for the 1000 records in the German credit data set. 
 Note that there are only 921 unique values. \textbf{Right:} The estimated probability of having a good credit risk 
 for the second 500 records in the German credit data set. The estimates are obtained from a multi-layer perceptron trained on the first 500 records in the German
  credit data set.}\label{plot_histogram_1D_data}
\end{figure}

\begin{table}[h]
\caption{Experiment on German credit data set. Clustering 1000 people according to their credit amount. Target cluster sizes $t_i=\frac{1000}{k}$, $i\in [k]$. 
\textsc{Naive}$=$naive clustering that matches the target cluster sizes, \textsc{DP}$=$dynamic programming approach of Section~\ref{section_1dim}, 
\textsc{$k$-means}$=k$-means initialized with medians of the 
clusters of the naive clustering, \textsc{$k$-me++}$=k$-means++. 
Results for 
\textsc{$k$-me++}  
averaged over 100 runs. Best values in~bold.
}
\label{experiment_1Da_part2}
\vskip 0.15in
\begin{center}
\begin{small}
\begin{sc}
\begin{tabular}{lccccccccccc}
\toprule
 & $\Nrunf$ & $\Maxviol$ & $\Obj$ & $\sqcost$ & $\cost$ && $\Nrunf$ & $\Maxviol$ & $\Obj$ & $\sqcost$ & $\cost$ \\
\midrule
&  \multicolumn{5}{c}{$k=10$} &&  \multicolumn{5}{c}{$k=20$}\\
\addlinespace[0.1cm]
Naive & 113 & 2.16 & \textbf{0} & 2.18 & 13.88 && 92 & 3.17 & \textbf{0} & 0.8 & 7.32\\
DP & \textbf{0} & \textbf{1.0} & 131 & 0.37 & 9.29 && \textbf{0} & \textbf{1.0} & 37 & 0.15 & 4.87\\
$k$-means~ & 4 & 1.01 & 136 & 0.37 & \textbf{8.91} && 5 & 1.01 & 37 & 0.28 & 5.74\\
$k$-me++~& 2.51 & 1.01 & 159.9 & \textbf{0.34} & 9.59 && 6.73 & 1.05 & 98.4 & \textbf{0.08} & \textbf{4.78}\\
\bottomrule
\end{tabular}
\end{sc}
\end{small}
\end{center}
\vskip -0.1in
\end{table}

 \begin{table}[h!]
\caption{Experiment on German credit data set. Clustering the second 500 people according to their estimated probability of having a good credit risk. 
Target cluster sizes $t_i=\frac{500}{k}$, $i\in [k]$. 
\textsc{Naive}$=$naive clustering that matches the target cluster sizes, \textsc{DP}$=$dynamic programming approach of Section~\ref{section_1dim}, 
\textsc{$k$-means}$=k$-means initialized with medians of the 
clusters of the naive clustering, \textsc{$k$-me++}$=k$-means++. 
Results for 
\textsc{$k$-me++}  
averaged over 100 runs. Best values in~bold.
}
\label{experiment_1Db}
\vskip 0.15in
\begin{center}
\begin{small}
\begin{sc}
\begin{tabular}{cclccccc}
\toprule
& ~~~~~Target cluster sizes~~~~~ & & $\Nrunf$ & $\Maxviol$ & $\Obj$ & $\sqcost$ & $\cost$ \\
\midrule
\parbox[t]{4mm}{\multirow{4}{*}{$k=5$}} & 
\multirow{4}{*}{\shortstack{$t_1=\ldots=t_5=100$}} 
& Naive & 197 & 58.28 & \textbf{0} & 6.8 & 16.91 \\
&&DP & \textbf{0} & \textbf{0.99} & 214 & 0.64 & 6.9 \\
&&$k$-means & 1 & 1.02 & 212 & 0.64 &  \textbf{6.85}\\
&&$k$-me++ & 0.71 & 1.01 & 220.66 & \textbf{0.63} &  7.0 \\
\midrule
\parbox[t]{4mm}{\multirow{4}{*}{$k=10$}} & \multirow{4}{*}{\shortstack{$t_1=\ldots=t_{10}=50$}} 
& Naive & 162 & 10.27 & \textbf{0} & 1.82 & 8.35 \\
&&DP & \textbf{0} & \textbf{0.98} & 217 & 0.19 & 3.36   \\
&&$k$-means & 5 & 1.06 & 207 & 0.37 &  4.3\\
&&$k$-me++ & 0.98 & 1.01 & 248.66 & \textbf{0.12} &  \textbf{3.13}\\
\midrule
\parbox[t]{4mm}{\multirow{4}{*}{$k=20$}} & \multirow{4}{*}{\shortstack{$t_1=\ldots=t_{20}=25$}} 
& Naive & 116 & 9.64 & \textbf{0} & 0.43 & 4.06 \\
&&DP & \textbf{0} & \textbf{1.0} & 155 & 0.17 & 2.96  \\
&&$k$-means & 33 & 2.13 & 95 & 0.1 &  2.16  \\
&&$k$-me++ &  2.62 & 1.06 & 239.64 & \textbf{0.03} &  \textbf{1.34}  \\ 
\midrule
\parbox[t]{4mm}{\multirow{4}{*}{$k=50$}} & \multirow{4}{*}{\shortstack{$t_1=\ldots=t_{50}=10$}} 
& Naive &  73 & 3.8 & \textbf{0} & 0.06 & 1.54 \\
&&DP & \textbf{0} & \textbf{1.0} & 24 & 0.04 & 1.32 \\
&&$k$-means &  28 & 2.39 & 13 & 0.04 & 1.28 \\
&&$k$-me++ &  3.07 & 1.24 & 234.17 & \textbf{0.0} &  \textbf{0.41}  \\
\bottomrule
\end{tabular}
\end{sc}
\end{small}
\end{center}
\vskip -0.1in

\vspace{4mm}
\end{table}

\vspace{6mm}
\section{
Addendum to Section~\ref{section_experiments_1D}
}\label{appendix_exp_1dim}

Figure~\ref{plot_histogram_1D_data} shows the histograms of the two  1-dimensional data sets 
that we 
used in the  experiments  of Section~\ref{section_experiments_1D}.
 
Table~\ref{experiment_1Da_part2} shows the results for the first experiment of Section~\ref{section_experiments_1D} when $k=10$ or $k=20$. 
 
Table~\ref{experiment_1Db} and Table~\ref{experiment_1Dc} provide the results for the second experiment of Section~\ref{section_experiments_1D}. 
In Table~\ref{experiment_1Db}, we 
consider 
uniform target cluster sizes $t_i=\frac{500}{k}$, $i\in [k]$, while in Table~\ref{experiment_1Dc} we consider various non-uniform target cluster sizes. The interpretation of the 
results is similar as for the first experiment of Section~\ref{section_experiments_1D}. Most notably, \textsc{$k$-means} can be quite unfair with up to 33 data points being 
treated unfair when $k$ is large, whereas \textsc{$k$-me++} produces very fair clusterings with not more than three data points being treated unfair. However, $\textsc{$k$-me++}$ 
performs 
very poorly
in terms of $\Obj$, which can be almost ten times as large as for \textsc{$k$-means} and our dynamic programming approach \textsc{DP} (cf. Table~\ref{experiment_1Db}, $k=50$). 

The MLP that we used 
for predicting the label (good vs. bad credit risk) in the second experiment of Section~\ref{section_experiments_1D} has 
three hidden layers of size 100, 50 and 20, respectively, and a test accuracy of 0.724.

 \begin{table}[h!]
\caption{Experiment on German credit data set. Clustering the second 500 people according to their estimated probability of having a good credit risk. 
Various 
non-uniform 
target cluster sizes.  
\textsc{Naive}$=$naive clustering that matches the target cluster sizes, \textsc{DP}$=$dynamic programming approach of Section~\ref{section_1dim}, \textsc{$k$-means}$=k$-means initialized with medians of the 
clusters of the naive clustering, \textsc{$k$-me++}$=k$-means++. 
Results for 
\textsc{$k$-me++}  
averaged over 100 runs. Best values in~bold.
}
\label{experiment_1Dc}
\vskip 0.15in
\begin{center}
\begin{small}
\begin{sc}
\begin{tabular}{cclccccc}
\toprule
& ~~~~~Target cluster sizes~~~~~ & & $\Nrunf$ & $\Maxviol$ & $\Obj$ & $\sqcost$ & $\cost$ \\
\midrule
\parbox[t]{9mm}{\multirow{4}{*}{$k=12$}} & \multirow{4}{*}{\shortstack{$t_i=\begin{cases}50 &\text{\textnormal{for}~}3\leq i\leq 10 \\25& \text{\textnormal{else}} \end{cases}$}} 
& Naive & 188 & 12.85 & \textbf{0} & 1.82 & 8.34 \\
&&DP & \textbf{0} & \textbf{0.97} & 232 & 0.17 & 3.06 \\ 
&&$k$-means & 3 & 1.05 & 217 & 0.18 &  3.18 \\
&&$k$-me++ & 1.25 & 1.03 & 255.1 & \textbf{0.08} &  \textbf{2.36} \\
\midrule
\parbox[t]{9mm}{\multirow{6}{*}{$k=12$}} & \multirow{6}{*}{\shortstack{$t_1=t_{12}=10$,\\$t_2=t_{11}=15$,\\$t_3=t_{10}=25$,\\$t_4=t_9=50$,\\$t_5=t_8=50$,\\$t_6=t_7=100$}} 
& & & & & &\\
&& Naive & 251 & 65.99 & \textbf{0} & 2.2 & 10.28 \\ 
&&DP & \textbf{0} & \textbf{0.97} & 247 & 0.17 & 3.06\\   
&&$k$-means & 5 & 1.16 & 247 & 0.14 &  2.64 \\
&&$k$-me++ & 1.22 & 1.03 & 270.5 & \textbf{0.08} &  \textbf{2.37} \\
& & & & & & & \\
\midrule
\parbox[t]{9mm}{\multirow{4}{*}{$k=20$}} & \multirow{4}{*}{\shortstack{$t_i=\begin{cases}10 &\text{\textnormal{for}~}i=1,3,5,\ldots \\40& \text{\textnormal{for}~}i=2,4,6,\ldots \end{cases}$}}
& Naive & 189 & 137.31 & \textbf{0} & 0.97 & 5.7 \\ 
&&DP & \textbf{0} & \textbf{1.0} & 140 & 0.17 & 2.96 \\ 
&&$k$-means & 30 & 1.91 & 91 & 0.09 &  2.13 \\
&&$k$-me++ & 2.37 & 1.07 & 225.17 & \textbf{0.03} &  \textbf{1.35} \\ 
\midrule
\parbox[t]{9mm}{\multirow{4}{*}{$k=20$}} & \multirow{4}{*}{\shortstack{$t_i=\begin{cases}115 &\text{\textnormal{for}~}i=10,11 \\15& \text{\textnormal{else}} \end{cases}$}}
& Naive & 224 & 215.88 & \textbf{0} & 1.96 & 9.11 \\ 
&&DP & \textbf{0} & \textbf{1.0} & 165 & 0.17 & 2.96 \\ 
&&$k$-means & 25 & 2.04 & 156 & 0.09 &  1.92 \\ 
&&$k$-me++ & 2.71 & 1.07 & 249.9 & \textbf{0.03} &  \textbf{1.34} \\
\bottomrule
\end{tabular}
\end{sc}
\end{small}
\end{center}
\vskip -0.1in

\vspace{1.4cm}
\end{table}



\vspace{6mm}
\section{
Addendum to Section~\ref{section_experiments_general}
}\label{appendix_exp_general}

In Appendix~\ref{example_group_fair_vs_indi_fair}, we present a simple example that shows that it really 
depends on the data set whether a group-fair clustering is individually fair or not.

In Appendix~\ref{example_local_search}, we provide an example illustrating why the local search idea outlined
in Section~\ref{section_experiments_general} does not work.

In Appendix~\ref{appendix_pruning_strategy}, we provide the pseudocode of our proposed heuristic  
to greedily prune a hierarchical clustering with the goal of minimizing $\Nrunf$ or $\Maxviol$.

\begin{figure*}[h]
\centering
%
%
%
%
\includegraphics[width=\textwidth]{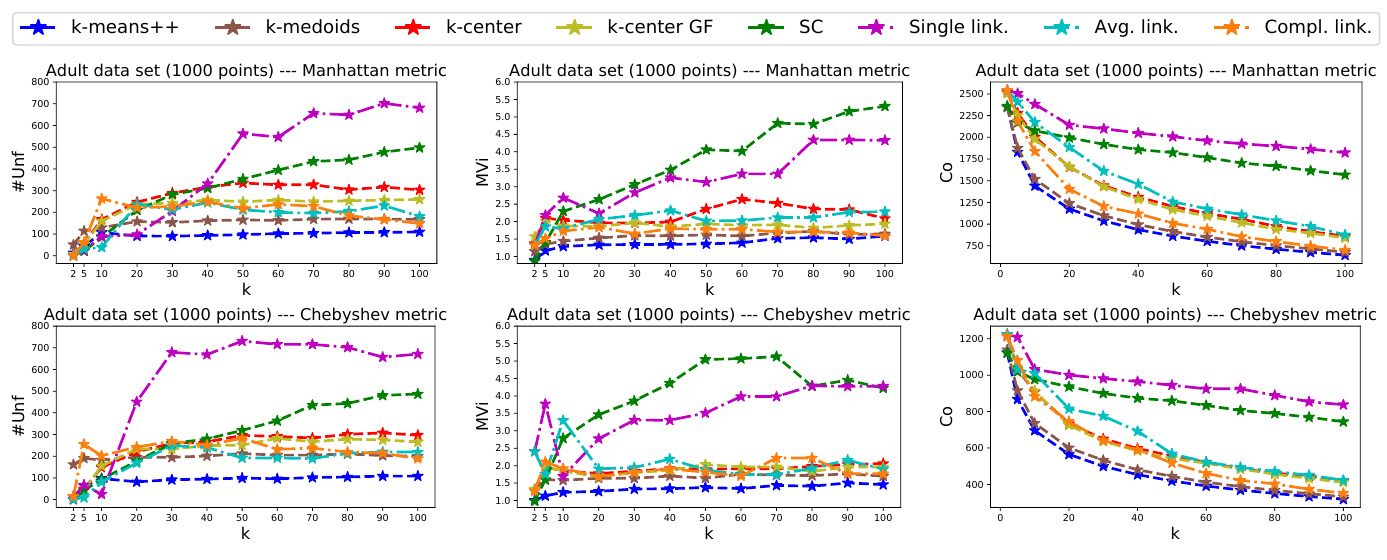}

\caption{Adult data set --- similar plots as in Figure~\ref{exp_gen_standard_alg_Adult}, but for the Manhattan \textbf{(top row)} and Chebyshev metric \textbf{(bottom row)}: 
$\Nrunf$ \textbf{(left)}, $\Maxviol$ 
\textbf{(middle)} and $\cost$ \textbf{(right)}
for the clusterings produced by the 
various 
standard 
algorithms as a function of the number of clusters~$k$.
}\label{exp_gen_standard_alg_Adult_2}
\end{figure*}

\begin{figure*}[h]
\centering
%
%
%
\includegraphics[width=\textwidth]{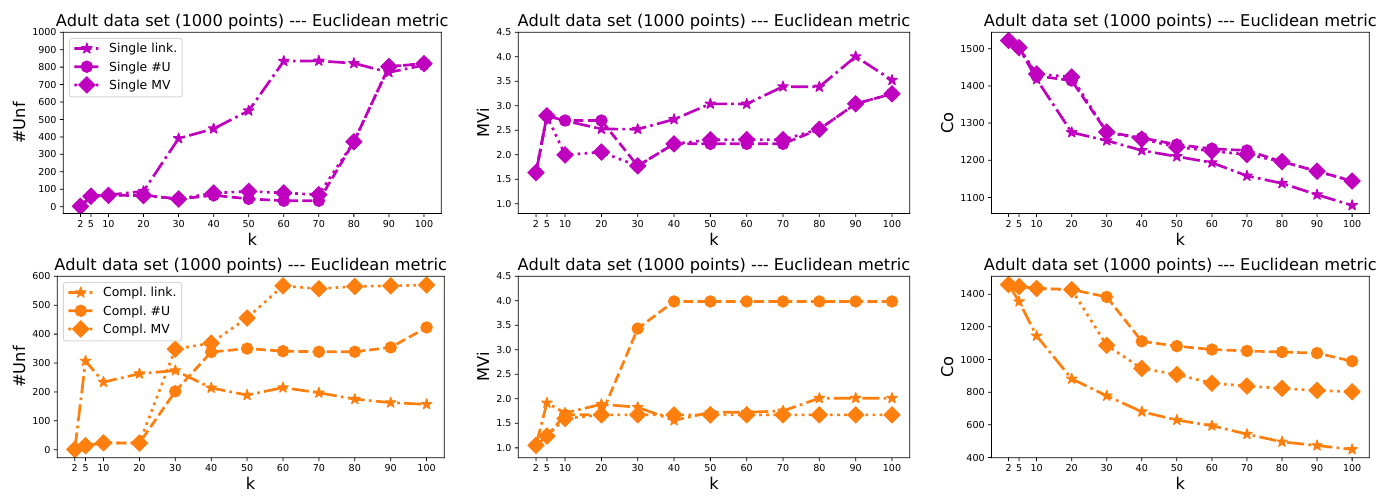}

\caption{Adult data set with Euclidean metric --- similar plots as in Figure~\ref{exp_gen_heuristics_Adult}, but for 
single \textbf{(top row)} and complete linkage clustering \textbf{(bottom row)}: $\Nrunf$ 
\textbf{(left)}, $\Maxviol$ 
\textbf{(middle)} and $\cost$ \textbf{(right)}
for the clusterings produced by single / complete linkage clustering
and the two variants of our heuristic 
approach 
to improve it: the first ($\#$U in the legend) greedily chooses splits as to minimize $\Nrunf$, the second 
(MV in the legend) 
as to minimize $\Maxviol$. 
}
\label{exp_gen_heuristics_Adult_appendix}
\end{figure*}

\begin{figure*}[h]
\centering
\includegraphics[width=\textwidth]{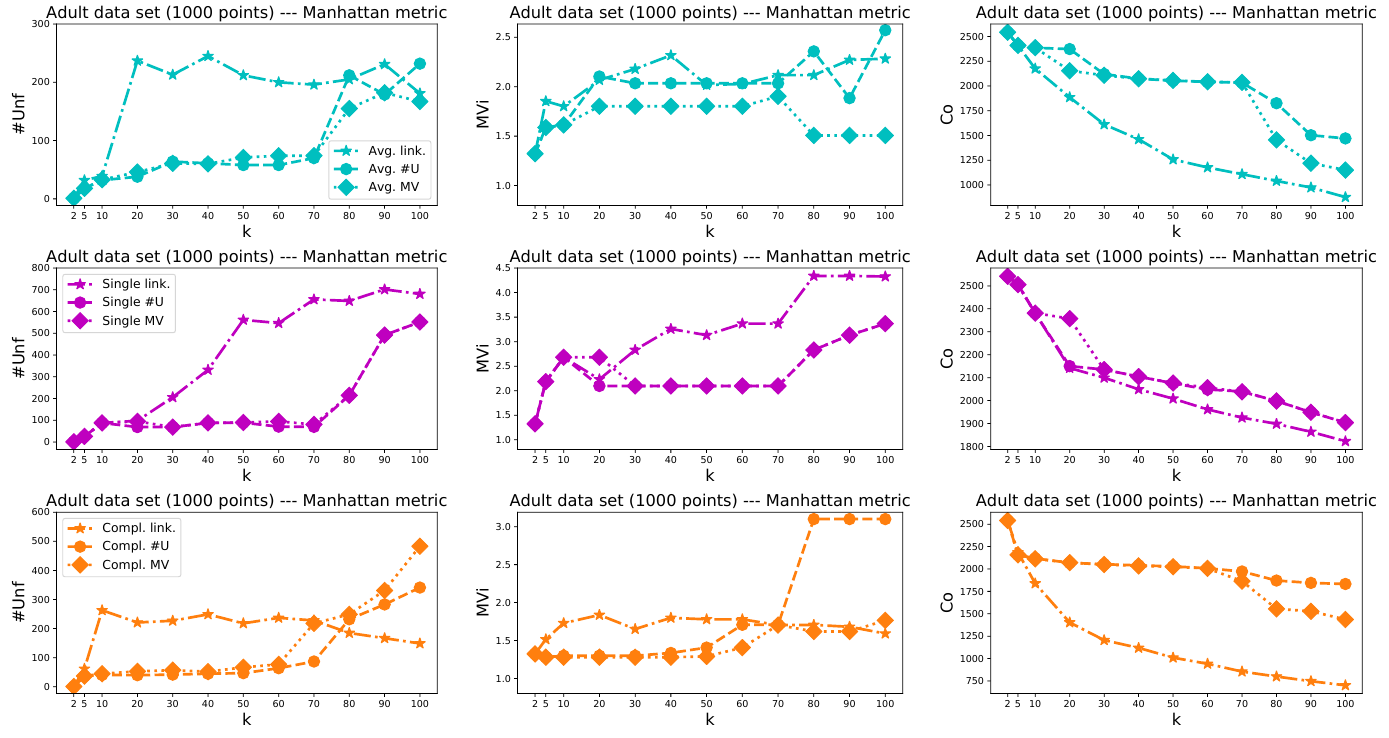}

\caption{Adult data set with Manhattan metric (similar plots as in Figures~\ref{exp_gen_heuristics_Adult} and~\ref{exp_gen_heuristics_Adult_appendix}, respectively): 
$\Nrunf$ 
\textbf{(left)}, $\Maxviol$ 
\textbf{(middle)} and $\cost$ \textbf{(right)}
for the clusterings produced by average  \textbf{(top row)} / single  \textbf{(middle row)} / complete linkage clustering \textbf{(bottom row)} 
and the two variants of our heuristic 
approach 
to improve it.
}
\label{exp_gen_heuristics_Adult_appendix_cityblock}
\end{figure*}

\begin{figure*}[h!]
\centering
\includegraphics[width=\textwidth]{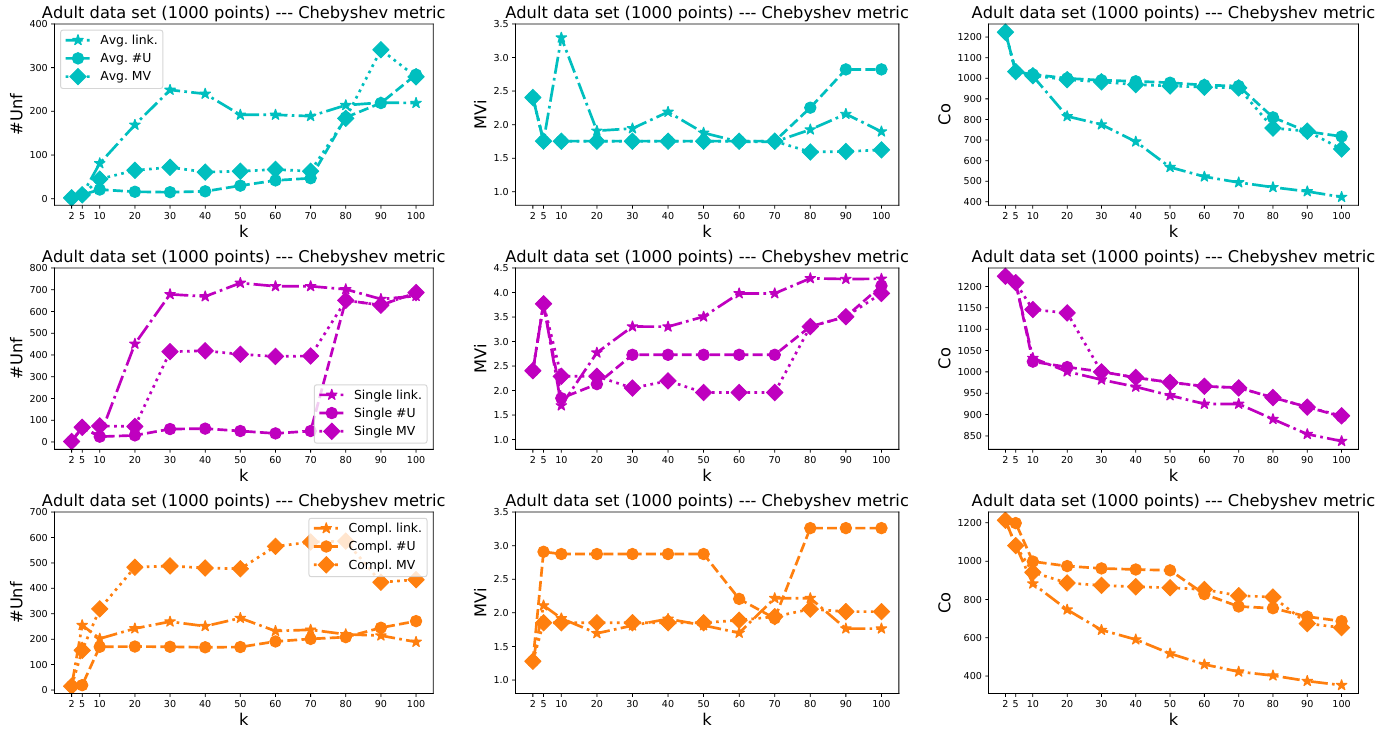}

\caption{Adult data set with Chebyshev metric (similar plots as in Figures~\ref{exp_gen_heuristics_Adult} and~\ref{exp_gen_heuristics_Adult_appendix}, respectively): 
$\Nrunf$ 
\textbf{(left)}, $\Maxviol$ 
\textbf{(middle)} and $\cost$ \textbf{(right)}
for the clusterings produced by average  \textbf{(top row)} / single  \textbf{(middle row)} / complete linkage clustering \textbf{(bottom row)} 
and the two variants of our heuristic 
approach 
to improve it.
}
\label{exp_gen_heuristics_Adult_appendix_chebyshev}
\end{figure*}

In Appendix~\ref{appendix_exp_general_adult}, we present the missing plots of Section~\ref{section_experiments_general} for the Adult data set: 
Figure~\ref{exp_gen_standard_alg_Adult_2} is analogous to Figure~\ref{exp_gen_standard_alg_Adult}, but for the Manhattan and Chebyshev metric, 
and shows $\Nrunf$, $\Maxviol$ and $\cost$ as a function of the number of clusters~$k$ for the various standard clustering algorithms. 
The results are very similar to the case of $d$ equaling the 
Euclidean metric (shown in Figure~\ref{exp_gen_standard_alg_Adult}),
and their interpretation is the same. Figure~\ref{exp_gen_heuristics_Adult_appendix} is analogous to Figure~\ref{exp_gen_heuristics_Adult}, but with single and complete linkage clustering 
instead of average linkage clustering. Just as for average linkage clustering (shown in Figure~\ref{exp_gen_heuristics_Adult}), 
we see that our heuristic approach can lead to a significant improvement in $\Nrunf$ 
(for complete linkage clustering, this is only true for $k\leq 20$, however) 
and also to some 
improvement in $\Maxviol$, but comes at the price of an increase in the clustering cost $\cost$. 
In Figures~\ref{exp_gen_heuristics_Adult_appendix_cityblock} and~\ref{exp_gen_heuristics_Adult_appendix_chebyshev} we study average / single / complete linkage clustering 
when $d$ equals the 
Manhattan or Chebyshev metric and 
make similar observations.

In Appendix~\ref{appendix_exp_general_drug}, we show the same set of experiments as in Figures~\ref{exp_gen_standard_alg_Adult} to~\ref{exp_gen_heuristics_Adult} and 
Figures~\ref{exp_gen_standard_alg_Adult_2} to~\ref{exp_gen_heuristics_Adult_appendix_chebyshev}, respectively, on the 
Drug Consumption data set. We used all 1885 records 
in the 
data set, and we 
used all 12 features 
describing 
a record (e.g., age, gender, or education), 
but did not use the information about the drug consumption of a 
record (this information is usually used as label when setting up a classification problem on the data set). We normalized the features to zero mean and unit variance. When 
running  
the standard clustering algorithms on the data set, we refrained  from running spectral clustering since the Scikit-learn implementation occasionally 
was not able to do the eigenvector computations and aborted with  a LinAlgError. 
Other than that, all results are largely consistent with the results for the Adult data set.

In Appendix~\ref{appendix_exp_general_liver}, we show the same set of experiments on the Indian Liver Patient data set. Removing four records with missing values, we ended up with 579 records, 
for which we used all 11 available features (e.g., age, gender, or total proteins). We normalized the features to zero mean and unit variance. 
Again, all results are largely consistent with the results for the Adult data set.

\subsection{Compatibility of Group Fairness and Individual Fairness}\label{example_group_fair_vs_indi_fair}

By means of a simple example we want to illustrate that it really depends on the data set whether group fairness and individual
fairness are compatible or at odds with each other. Here we consider the prominent 
group fairness notion for clustering of \citet{fair_clustering_Nips2017}, which 
asks that in each cluster, every demographic group is approximately equally represented. Let us assume that the data set
consists of the four 1-dimensional points 0, 1, 7 and 8 and the distance function $d$ is the ordinary Euclidean metric.
It is easy to see that the only individually fair 2-clustering is $\mathcal{C}=(\{0,1\}, \{7,8\})$. Now if there are two demographic 
groups $G_1$ and $G_2$ with $G_1=\{0,7\}$ and $G_2=\{1,8\}$, the clustering~$\mathcal{C}$ is perfectly fair according to the notion 
of \citeauthor{fair_clustering_Nips2017}. But if $G_1=\{0,1\}$ and $G_2=\{7,8\}$, the clustering~$\mathcal{C}$
is totally unfair according to the latter notion.

\subsection{Why Local Search Does not Work}\label{example_local_search}

\begin{figure}[t]
\centering
\includegraphics[scale=0.4]{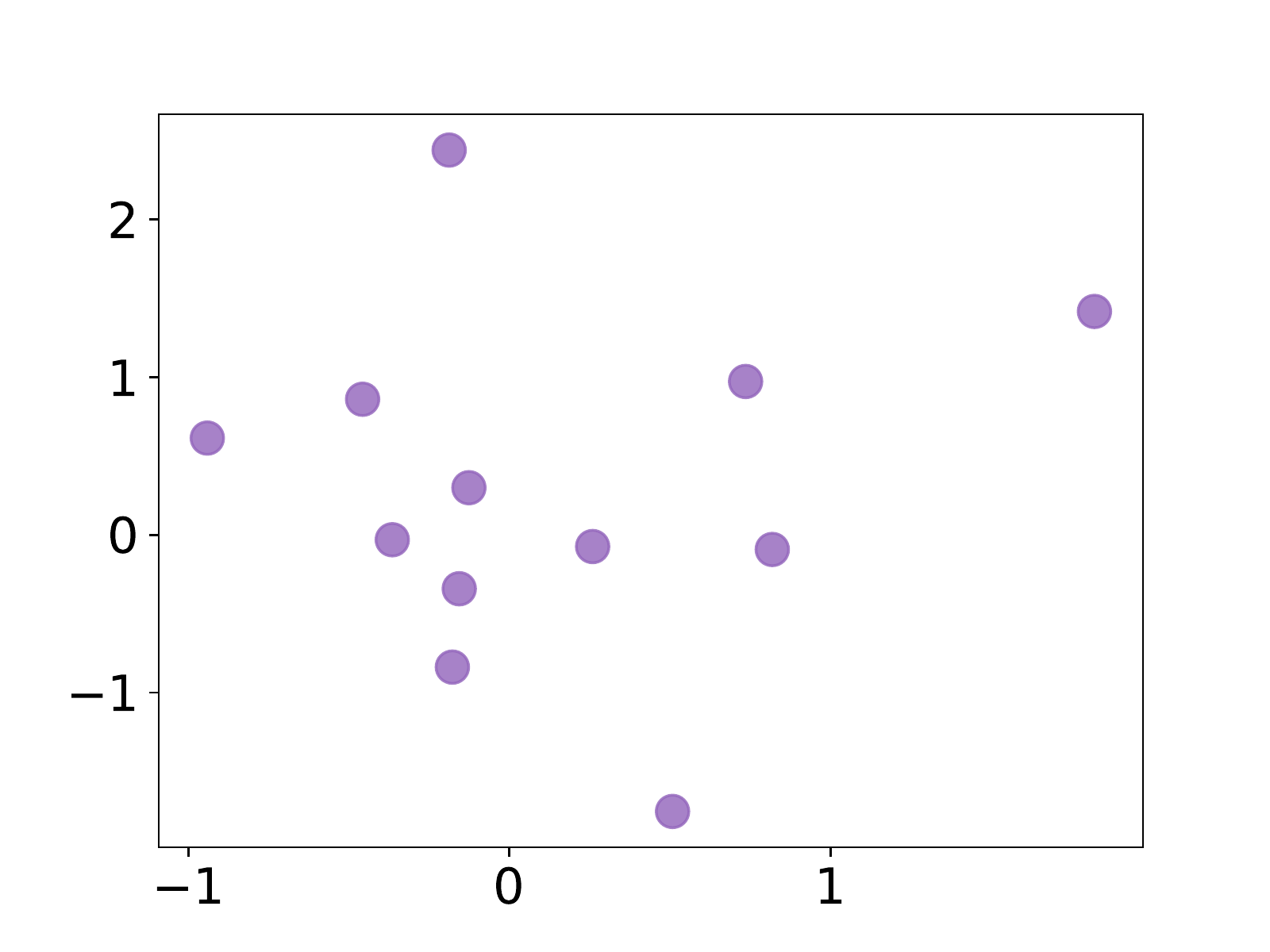}
\includegraphics[scale=0.4]{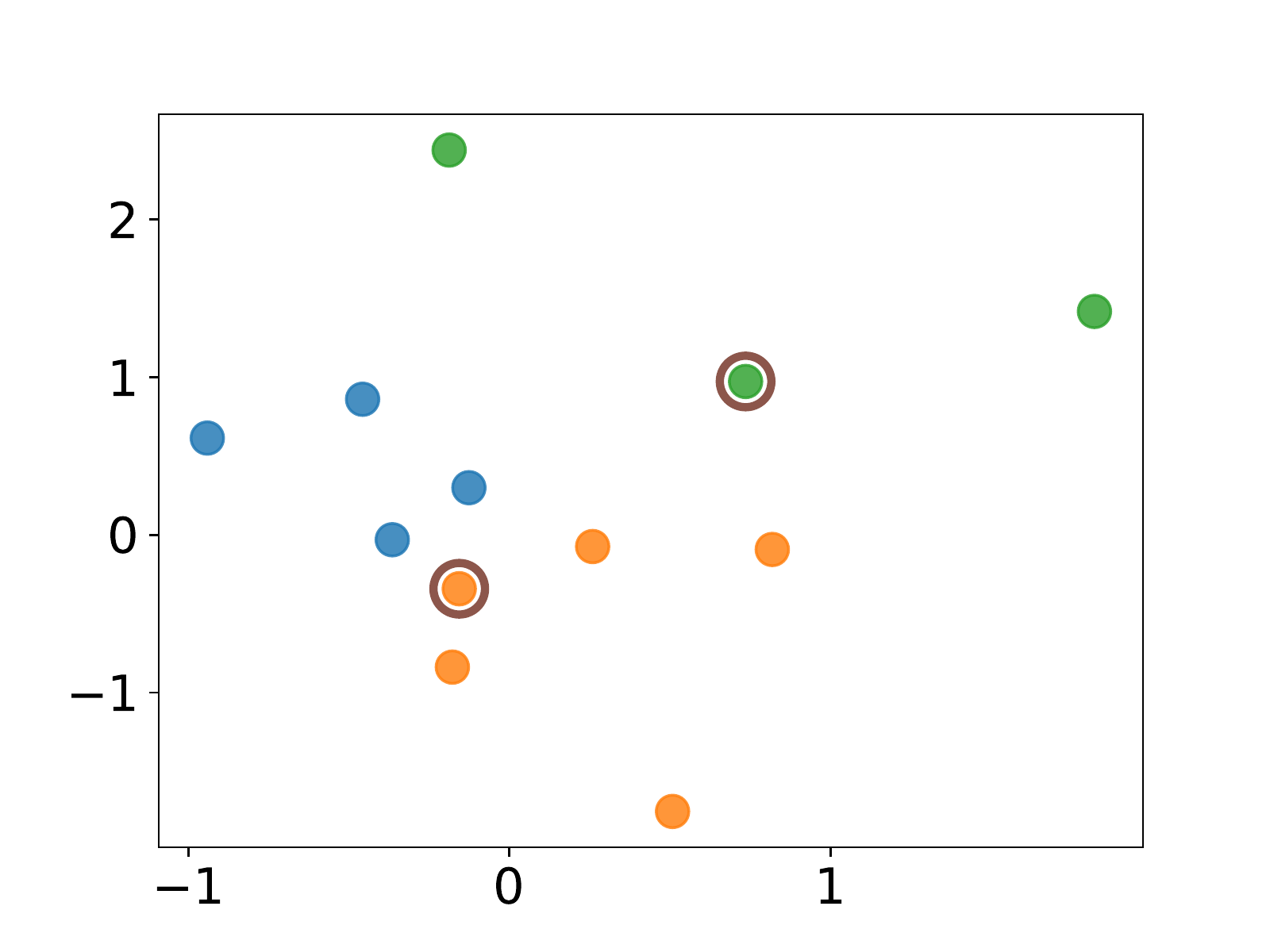}

\includegraphics[scale=0.4]{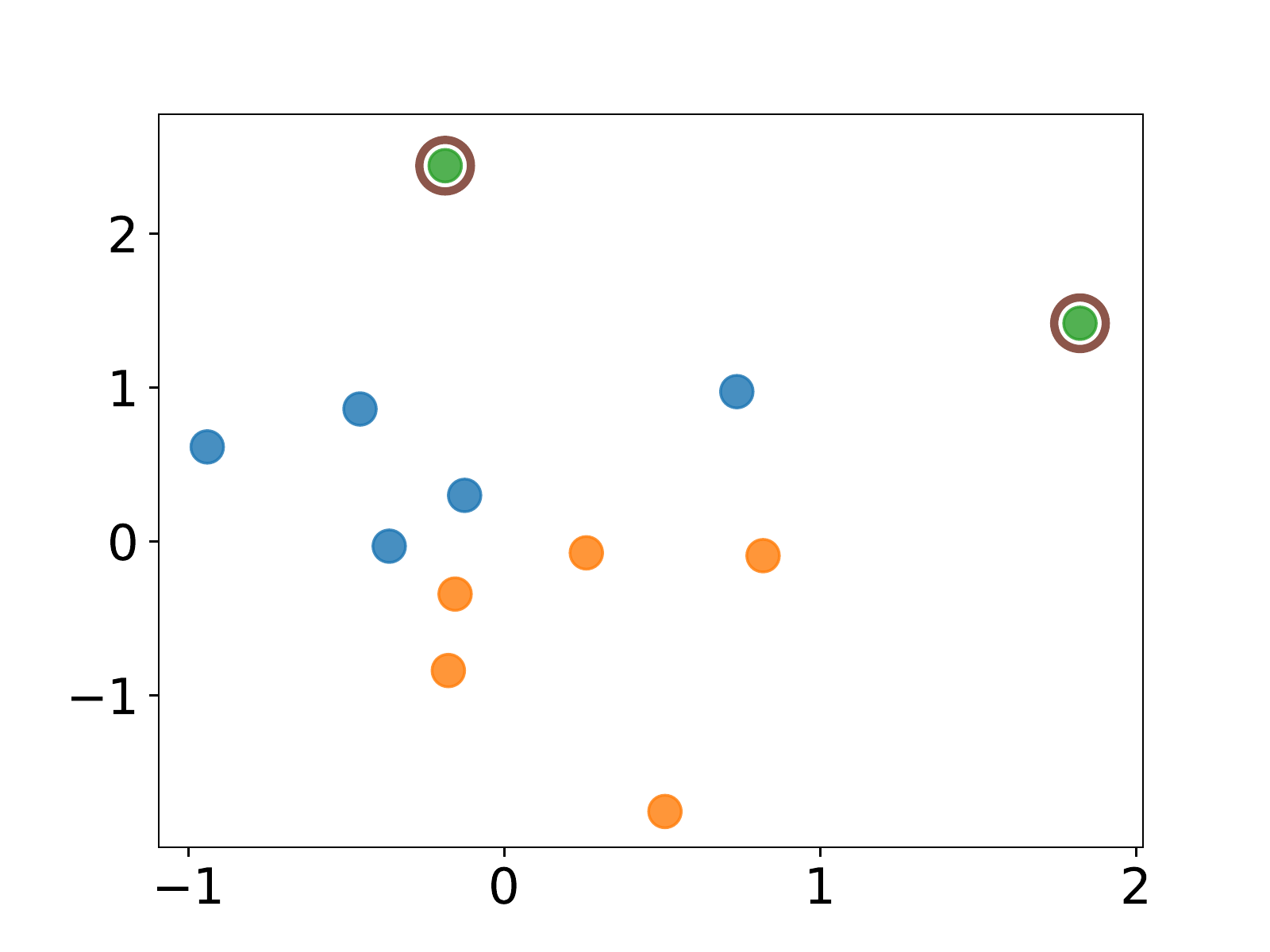}
\includegraphics[scale=0.4]{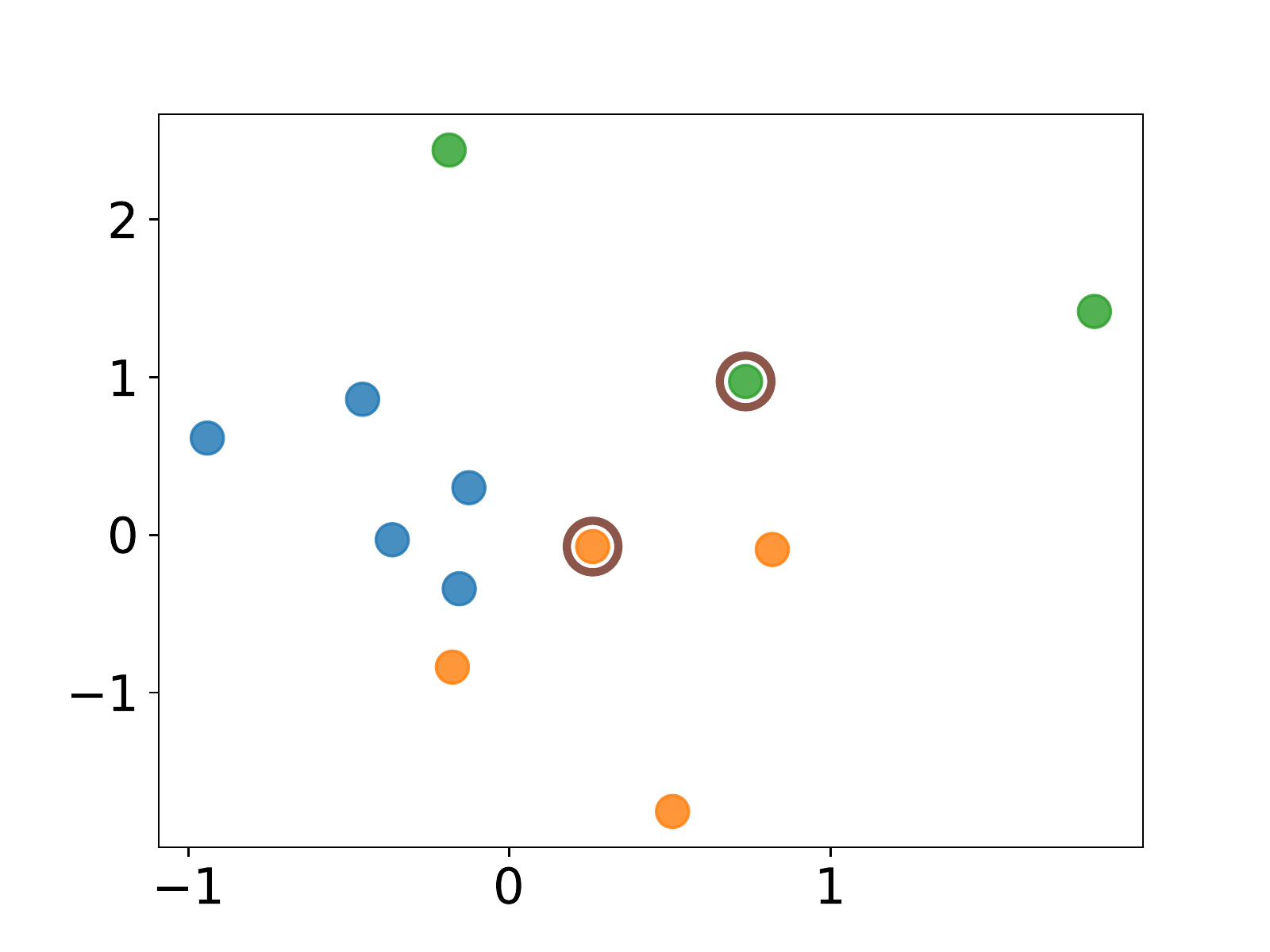}

\caption{An example illustrating why the local search idea outlined in Section~\ref{section_experiments_general} does not work.
\textbf{Top left:} 12 points in $\R^2$. 
\textbf{Top right:} A $k$-means clustering of the 12 points (encoded by color) 
with 
two points that are not treated individually fair (surrounded by a circle). 
\textbf{Bottom row:} After assigning one of the two points that are not treated fair in the $k$-means clustering to its closest cluster, that point is treated fair. However, 
now some points are treated unfair that were initially treated fair.}\label{fig_counterexample_local_search}
\end{figure}

Figure~\ref{fig_counterexample_local_search} presents an example illustrating why the local search idea outlined in 
Section~\ref{section_experiments_general} does not work: assigning a data point that is not treated fair to its closest cluster (so that that data point is treated fair) 
may cause other data points that are initially treated fair to be treated unfair 
after the reassignment.

\subsection{Pseudocode of our Proposed Heuristic Approach}\label{appendix_pruning_strategy}

Algorithm~\ref{pseudocode_pruning_strategy} provides the pseudocode of our proposed strategy 
to greedily prune a hierarchical clustering with the goal of minimizing $\Nrunf$ or $\Maxviol$.

\begin{algorithm}[H]
   \caption{Algorithm to greedily prune a hierarchical clustering}
   \label{pseudocode_pruning_strategy}
   
\begin{algorithmic}[1]
\vspace{1mm}
     \STATE {\bfseries Input:}  
       binary tree $T$ representing a hierarchical clustering obtained from running a linkage clustering algorithm; number of clusters~$k\in\{2,\ldots,|\dataset|\}$; 
       measure $meas\in\{\Nrunf,\Maxviol\}$ that one aims to optimize~for

\vspace{1mm}  
   \STATE {\bfseries Output:}
   a $k$-clustering $\mathcal{C}$
   
\vspace{3mm}
\STATE{\emph{\# Conventions:} 
\vspace{0mm}
\begin{itemize}[leftmargin=*]
\setlength{\itemsep}{1pt}
 \item \emph{for a node $v\in T$, we denote the left child of $v$ by $Left(v)$ and the right child by $Right(v)$}
\item 
\emph{for a $j$-clustering~$\mathcal{C}'=(C_1,C_2,\ldots,C_j)$, a cluster $C_l$ and $A,B\subseteq C_l$ with $A\dot{\cup} B =C_l$ we write 
$\mathcal{C}'|_{C_l\hookrightarrow A,B}$ for the $(j+1)-$clustering 
that we obtain by replacing the cluster $C_l$ with two clusters $A$ and $B$ in $\mathcal{C}'$}
\end{itemize}
}   
  
\vspace{4mm}
\STATE{Let $r$ be the root of $T$ and initialize the clustering $\mathcal{C}$ as $\mathcal{C}=(Left(r),Right(r))$}
\vspace{2pt}
\FOR{$i=1$ {\bfseries to} $k-2$ \textbf{by} $1$} 
\vspace{2pt}
\STATE{Set $$v^\star=\argmin_{v: v\text{ is a cluster in $\mathcal{C}$ with }|v|>1} meas(\mathcal{C}|_{v\hookrightarrow Left(v),Right(v)})$$ 
and
$$\mathcal{C}=\mathcal{C}|_{v^\star\hookrightarrow Left(v^\star),Right(v^\star)}$$
}

\vspace{4pt}
\ENDFOR

\RETURN~~$\mathcal{C}$
\end{algorithmic}
\end{algorithm}

\clearpage
\subsection{Adult Data Set}\label{appendix_exp_general_adult}

\vspace{1cm}

\vspace{2cm}

\clearpage

\vspace{2mm}

\vspace{10mm}

\clearpage
\subsection{Drug Consumption Data Set}\label{appendix_exp_general_drug}

\vspace{4mm}

\begin{figure*}[h]
\centering
%
%
%
%
%
%
%
%
\includegraphics[width=\textwidth]{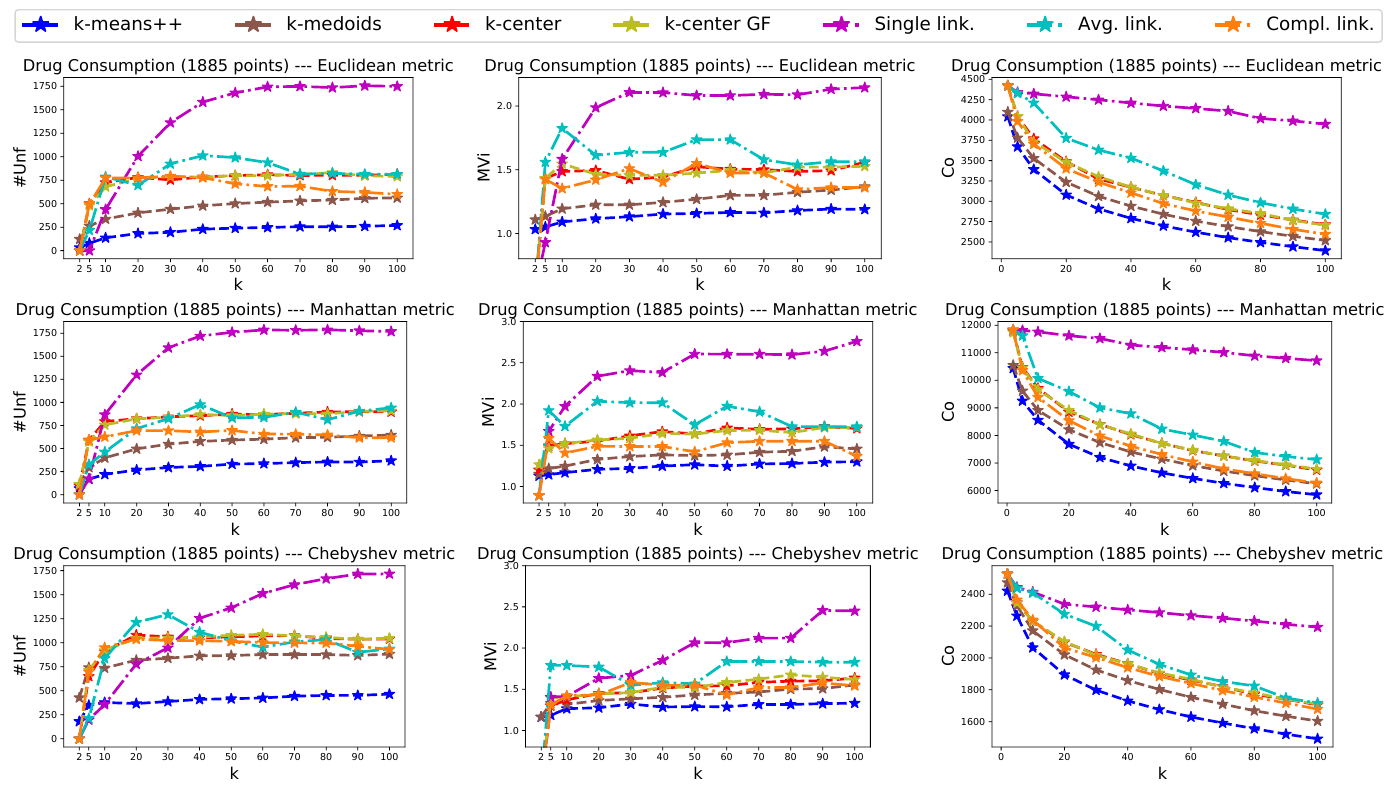}

\caption{Drug Consumption data set: $\Nrunf$ 
\textbf{(left)}, $\Maxviol$ 
\textbf{(middle)} and $\cost$ \textbf{(right)} 
for the clusterings produced by the 
various 
standard 
algorithms as a function of 
$k$ 
for the Euclidean \textbf{(top row)}, Manhattan \textbf{(middle row)} and Chebyshev metric \textbf{(bottom row)}.
}\label{exp_gen_standard_alg_DrugData}
\end{figure*}

\vspace{8mm}

\begin{figure*}[h!]
\centering
\includegraphics[width=\textwidth]{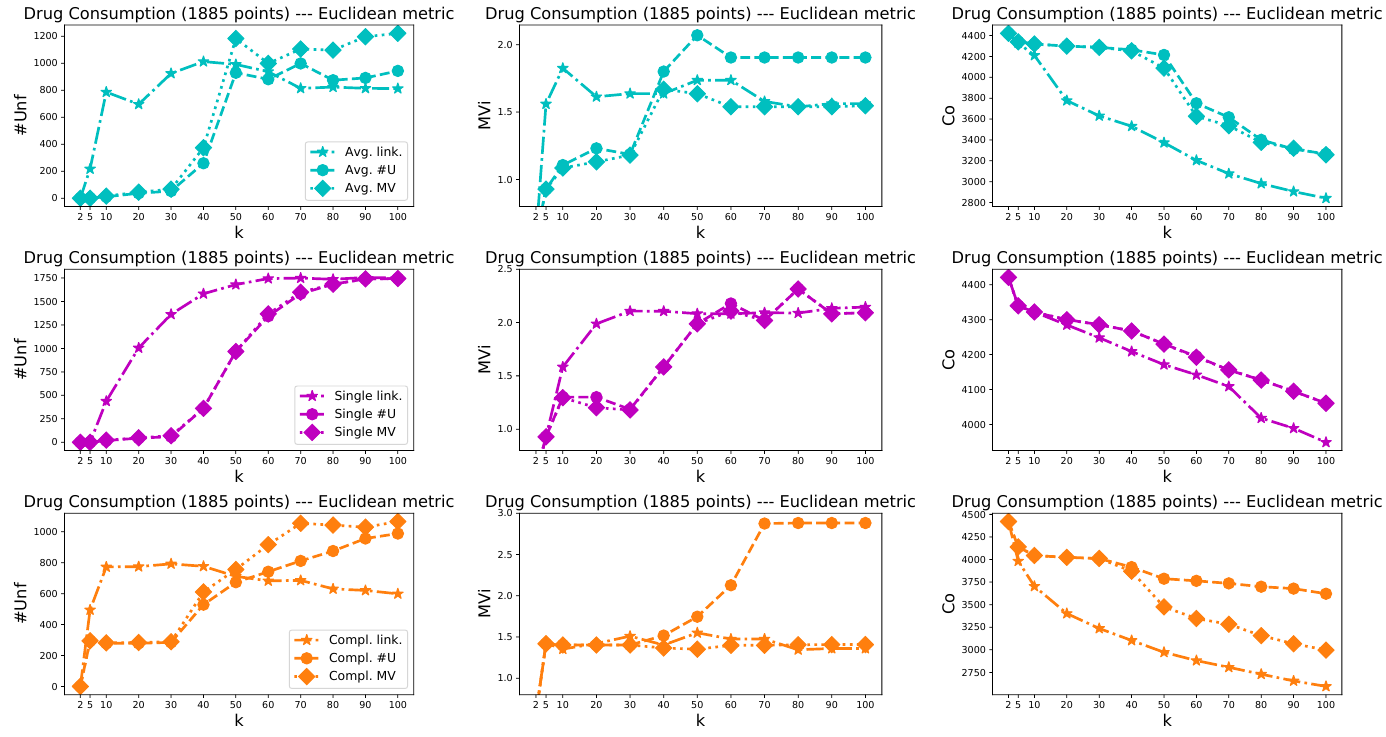}

\caption{Drug Consumption data set with Euclidean metric: $\Nrunf$ 
\textbf{(left)}, $\Maxviol$ 
\textbf{(middle)} and $\cost$ \textbf{(right)}
for the clusterings produced by 
average \textbf{(top row)} / 
single  \textbf{(middle row)} / complete linkage clustering \textbf{(bottom row)} 
and the two variants of our heuristic 
approach. 
}
\label{exp_gen_heuristics_DrugData_appendix}
\end{figure*}

\clearpage

\vspace{2mm}

\begin{figure*}[h]
\centering
\includegraphics[width=\textwidth]{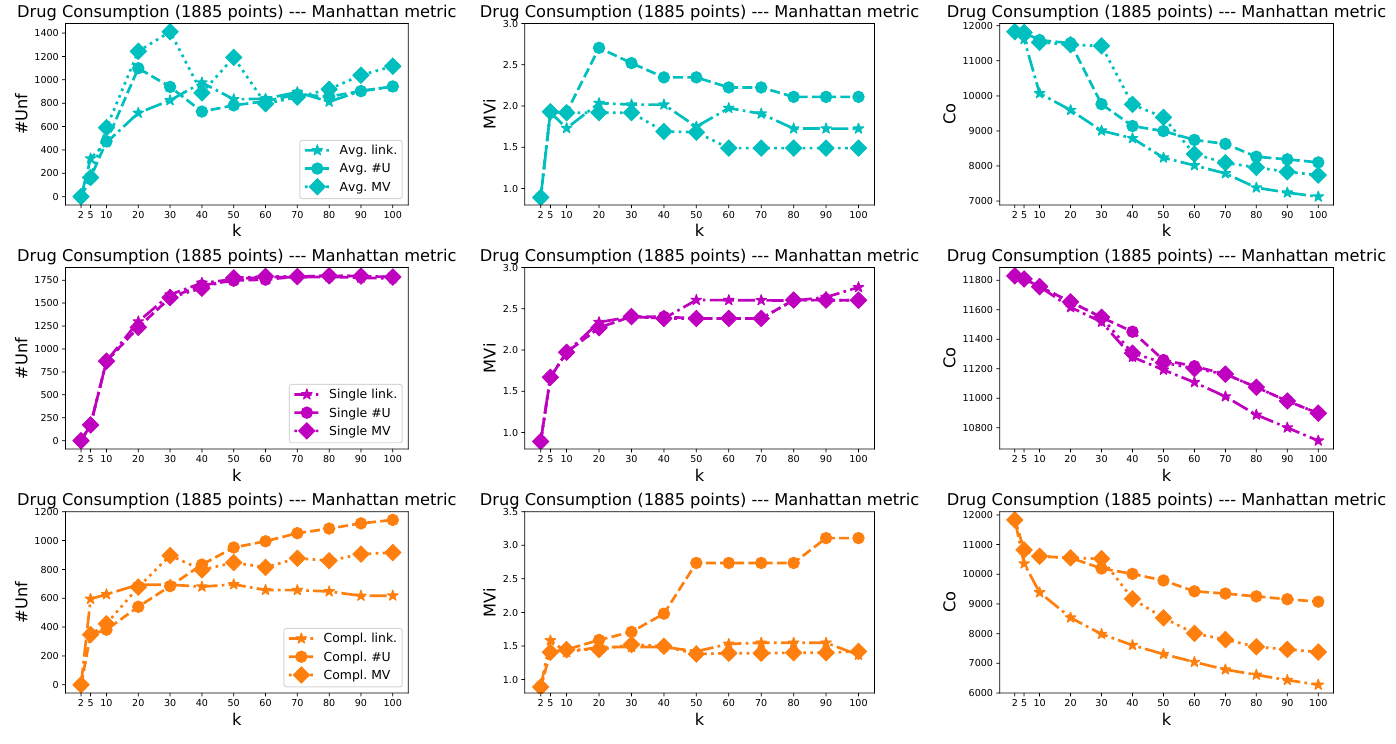}


\caption{Drug Consumption data set with Manhattan metric: $\Nrunf$ 
\textbf{(left)}, $\Maxviol$ 
\textbf{(middle)} and $\cost$ \textbf{(right)}
for the clusterings produced by 
average \textbf{(top row)} / 
single  \textbf{(middle row)} / complete linkage clustering \textbf{(bottom row)} 
and the two variants of our heuristic 
approach.}\label{exp_gen_heuristics_DrugData_appendix_cityblock}
\end{figure*}

\vspace{14mm}

\begin{figure*}[h!]
\centering
\includegraphics[width=\textwidth]{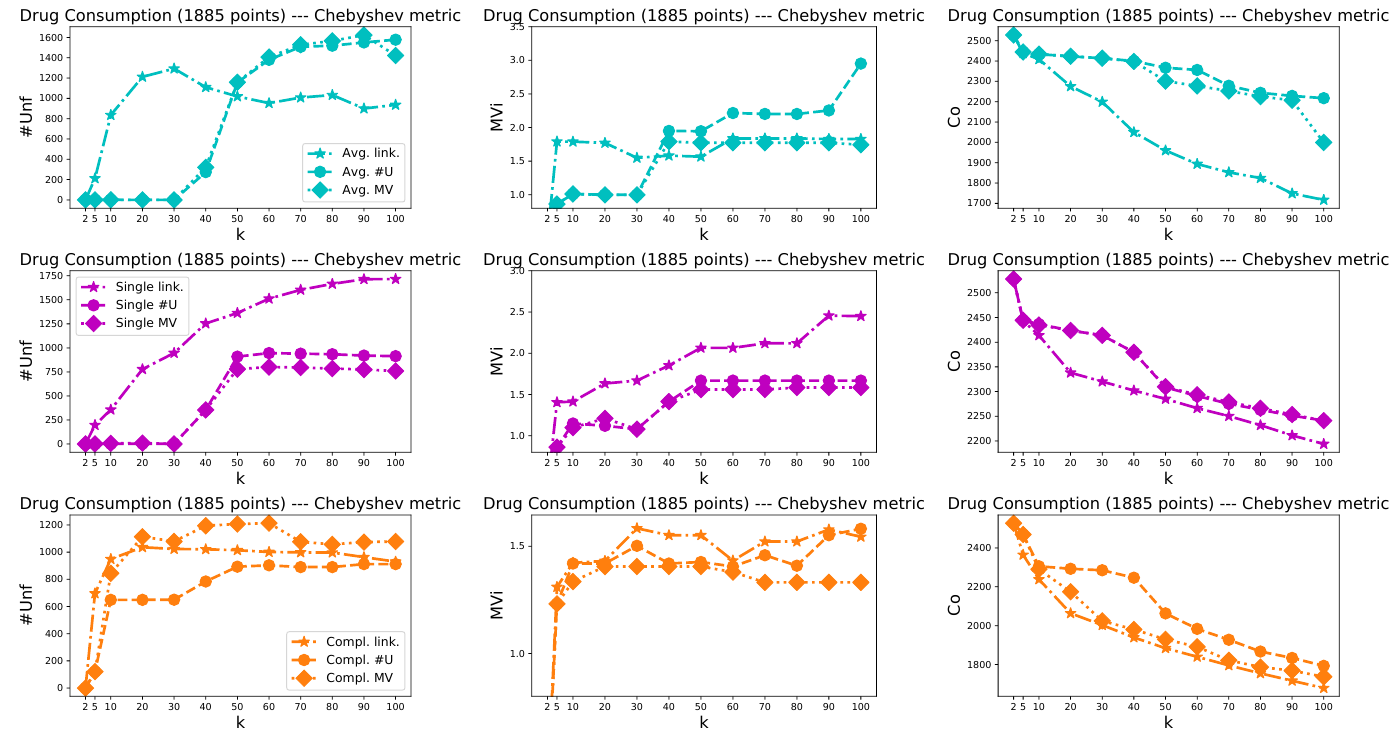}

\caption{Drug Consumption data set with Chebyshev metric: $\Nrunf$ 
\textbf{(left)}, $\Maxviol$ 
\textbf{(middle)} and $\cost$ \textbf{(right)}
for the clusterings produced by 
average \textbf{(top row)} / 
single  \textbf{(middle row)} / complete linkage clustering \textbf{(bottom row)} 
and the two variants of our heuristic 
approach.}\label{exp_gen_heuristics_DrugData_appendix_chebyshev}
\end{figure*}

\clearpage
\subsection{Indian Liver Patient Data Set}\label{appendix_exp_general_liver}

\vspace{4mm}

\begin{figure*}[h]
\centering
\includegraphics[width=\textwidth]{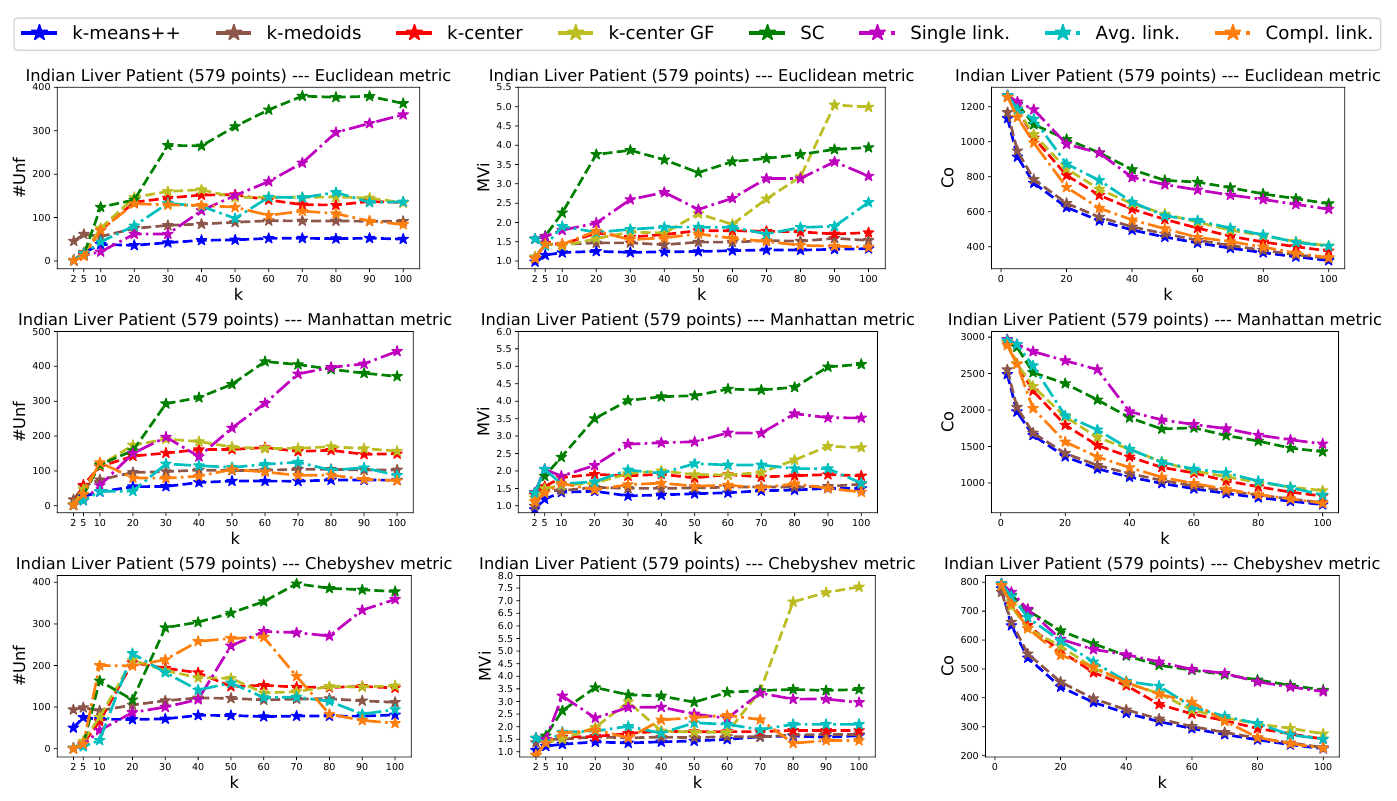}

\caption{Indian Liver Patient data set: $\Nrunf$ 
\textbf{(left)}, $\Maxviol$ 
\textbf{(middle)} and $\cost$ \textbf{(right)} 
for the clusterings produced by the 
various 
standard 
algorithms as a function of 
$k$ 
for the Euclidean \textbf{(top row)}, Manhattan \textbf{(middle row)} and Chebyshev metric \textbf{(bottom row)}.
}\label{exp_gen_standard_alg_IndianLiver}
\end{figure*}

\vspace{8mm}

\begin{figure*}[h!]
\centering
\includegraphics[width=\textwidth]{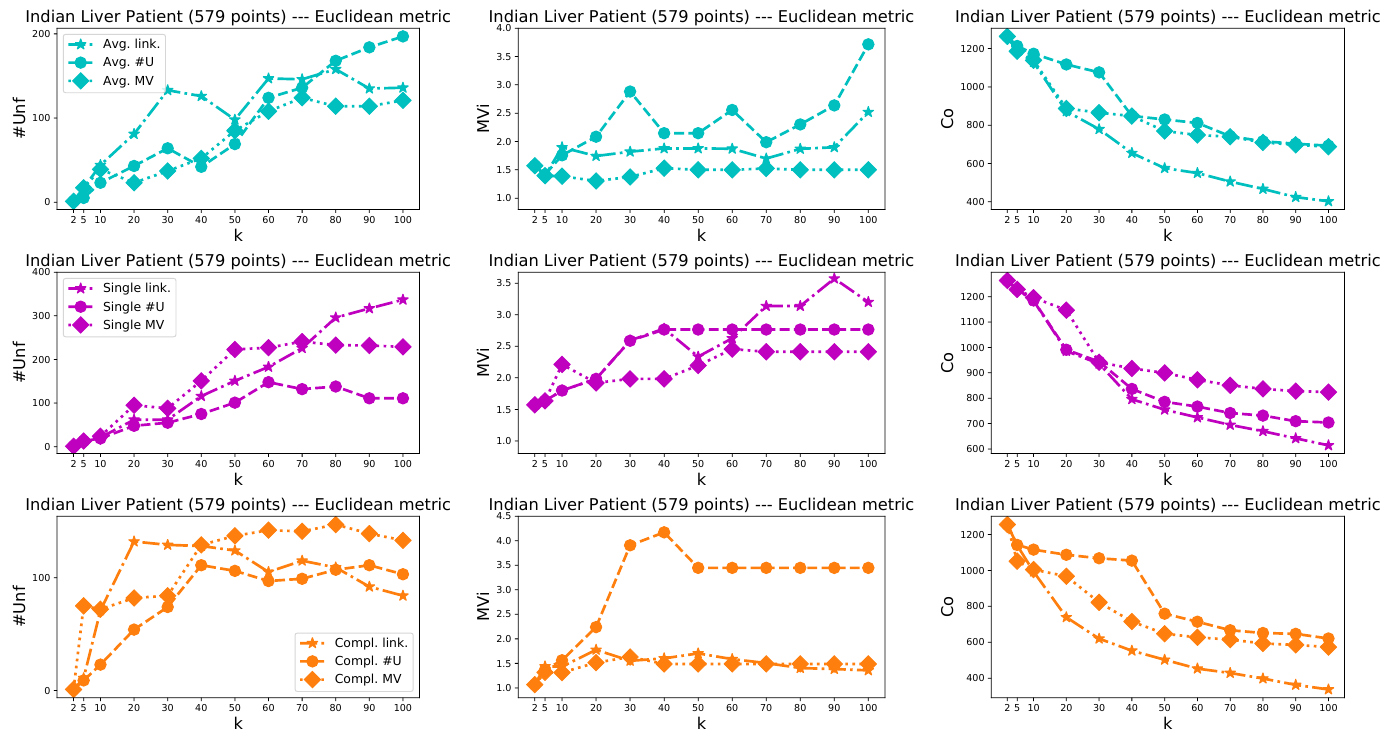}

\caption{Indian Liver Patient data set with Euclidean metric: $\Nrunf$ 
\textbf{(left)}, $\Maxviol$ 
\textbf{(middle)} and $\cost$ \textbf{(right)}
for the clusterings produced by 
average \textbf{(top row)} / 
single  \textbf{(middle row)} / complete linkage clustering \textbf{(bottom row)} 
and the two variants of our heuristic 
approach. 
}
\label{exp_gen_heuristics_IndianLiver_appendix}
\end{figure*}

\clearpage

\vspace{2mm}

\begin{figure*}[h]
\centering
\includegraphics[width=\textwidth]{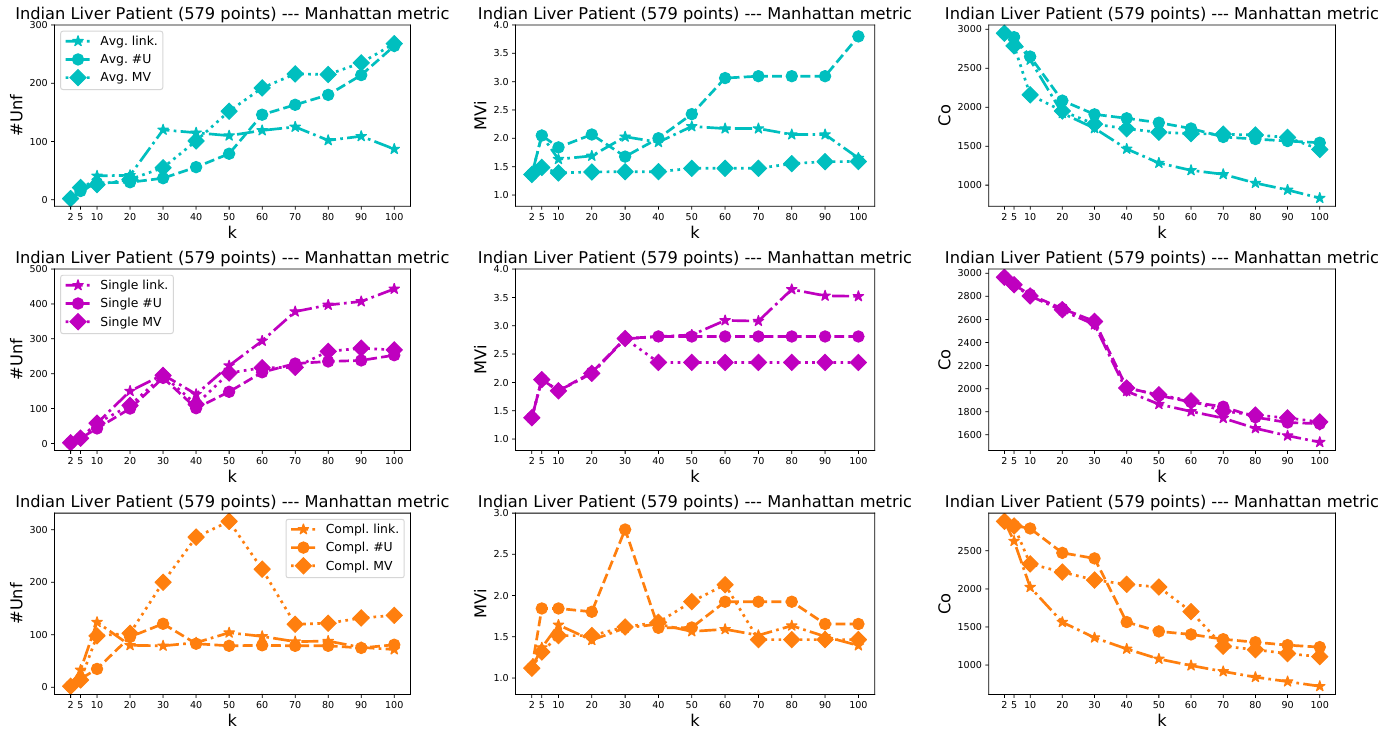}

\caption{Indian Liver Patient data set with Manhattan metric: $\Nrunf$ 
\textbf{(left)}, $\Maxviol$ 
\textbf{(middle)} and $\cost$ \textbf{(right)}
for the clusterings produced by 
average \textbf{(top row)} / 
single  \textbf{(middle row)} / complete linkage clustering \textbf{(bottom row)} 
and the two variants of our heuristic 
approach. 
}
\label{exp_gen_heuristics_IndianLiver_appendix_cityblock}
\end{figure*}

\vspace{14mm}

\begin{figure*}[h!]
\centering
\includegraphics[width=\textwidth]{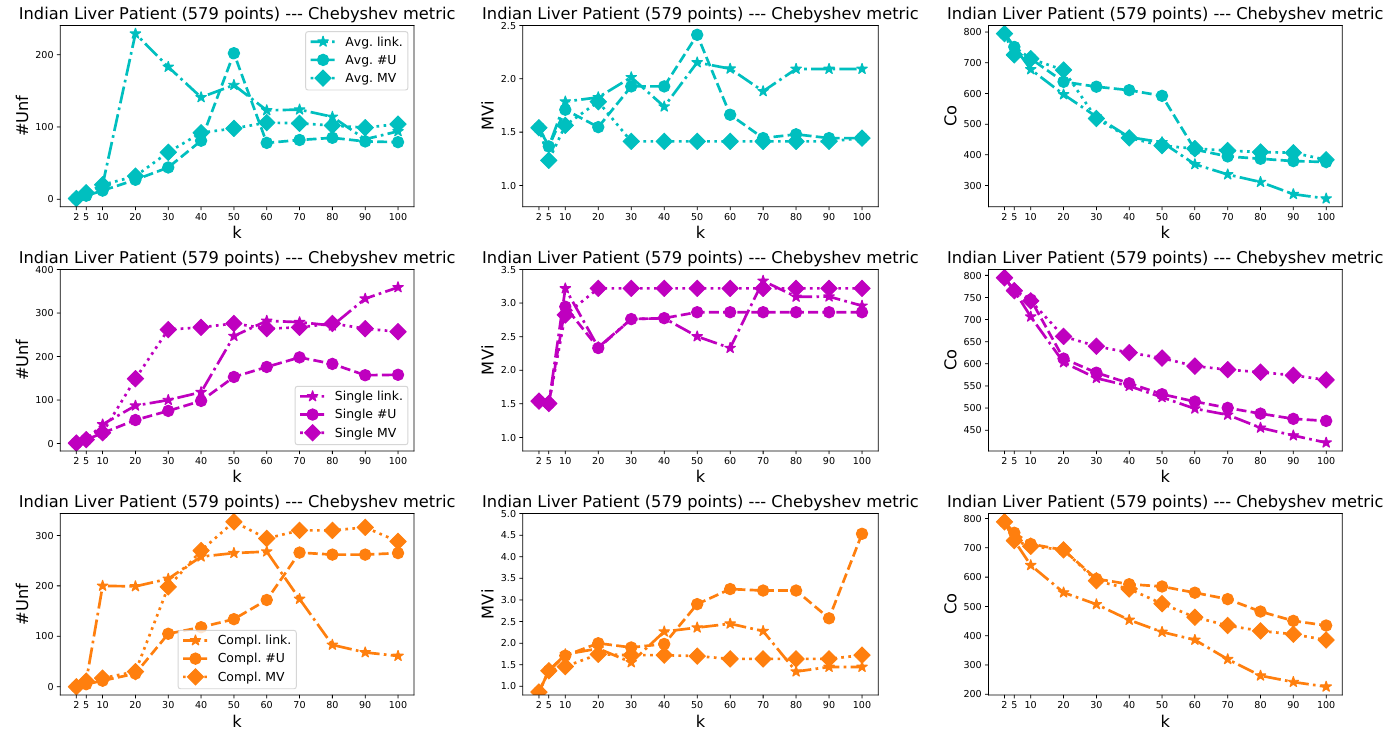}

\caption{Indian Liver Patient data set with Chebyshev metric: $\Nrunf$ 
\textbf{(left)}, $\Maxviol$ 
\textbf{(middle)} and $\cost$ \textbf{(right)}
for the clusterings produced by 
average \textbf{(top row)} / 
single  \textbf{(middle row)} / complete linkage clustering \textbf{(bottom row)} 
and the two variants of our heuristic 
approach. 
}
\label{exp_gen_heuristics_IndianLiver_appendix_chebyshev}
\end{figure*}

\end{document}